\newif\iffinal
\finaltrue

\documentclass[sigplan,screen]{acmart}
\usepackage[normalem]{ulem}

\usepackage{microtype}
\usepackage{xcolor}
\usepackage{hyperref}
\usepackage{url}
\usepackage{subfig}
\usepackage{multirow}
\usepackage{booktabs}
\usepackage{graphicx}
\usepackage{xspace}
\usepackage{algorithm}
\usepackage[noend]{algpseudocode}
\usepackage{balance}
\usepackage{amsmath}

\usepackage{amssymb}
\usepackage{amsthm}
\usepackage{threeparttable}
\usepackage{cleveref}
\usepackage{mathtools}

\newcommand{\Sys}{SpecInfer\xspace}
\newcommand{\sys}{SpecInfer\xspace}
\newcommand{\llama}{LLaMA\xspace}
\newcommand{\m}{\mathcal}

\newcommand{\ZJ}[1]{\textcolor{red}{ZJ: #1}}

\newcommand{\XM}[1]{\textcolor{blue}{XM: #1}}

\algnewcommand{\LeftComment}[1]{\Statex \(\triangleright\) #1}
\algnewcommand{\LineComment}[1]{\State \textcolor{violet}{\(\triangleright\) #1}}
\newcommand{\rev}[1]{{#1}}
\iffinal
\else 
\fi

\DeclareMathAlphabet{\mathcal}{OMS}{cmsy}{m}{n}

\usepackage{listings}
\definecolor{mygray}{RGB}{230,230,230}
\theoremstyle{plain}
\newtheorem{theorem}{Theorem}[section]

\theoremstyle{definition}
\newtheorem{definition}[theorem]{Definition}

\theoremstyle{remark}

\copyrightyear{2024} 
\acmYear{2024} 
\setcopyright{rightsretained} 
\acmConference[ASPLOS '24]{29th ACM International Conference on Architectural Support for Programming Languages and Operating Systems, Volume 3}{April 27-May 1, 2024}{La Jolla, CA, USA}
\acmBooktitle{29th ACM International Conference on Architectural Support for Programming Languages and Operating Systems, Volume 3 (ASPLOS '24), April 27-May 1, 2024, La Jolla, CA, USA}
\acmDOI{10.1145/3620666.3651335}
\acmISBN{979-8-4007-0386-7/24/04}

\begin{document}

\title[\Sys]{
\Sys: Accelerating Large Language Model Serving with Tree-based Speculative Inference and Verification}

\author{Xupeng Miao}\authornote{Equal contribution.} %
\affiliation{\institution{Carnegie Mellon University}\city{Pittsburgh}\state{PA}\country{USA}}
\orcid{0000-0002-9371-8358}
\email{xupeng@cmu.edu}

\author{Gabriele Oliaro}\authornotemark[1]
\affiliation{\institution{Carnegie Mellon University}\city{Pittsburgh}\state{PA}\country{USA}}
\orcid{0000-0001-5406-0736}
\email{goliaro@cs.cmu.edu}

\author{Zhihao Zhang}\authornotemark[1]
\affiliation{\institution{Carnegie Mellon University}\city{Pittsburgh}\state{PA}\country{USA}}
\orcid{0009-0002-8409-2717}
\email{zhihaoz3@andrew.cmu.edu}

\author{Xinhao Cheng}\authornotemark[1]
\affiliation{\institution{Carnegie Mellon University}\city{Pittsburgh}\state{PA}\country{USA}}
\orcid{0009-0009-3375-497X}
\email{xinhaoc@andrew.cmu.edu}

\author{Zeyu Wang}
\affiliation{\institution{Carnegie Mellon University}\city{Pittsburgh}\state{PA}\country{USA}}
\orcid{0009-0002-5756-2744}
\email{zeyuwang@alumni.cmu.edu}

\author{Zhengxin Zhang}
\affiliation{\institution{Tsinghua University}\city{Beijing}\country{China}}
\orcid{0000-0003-1578-2597}
\email{zhang-zx21@mails.tsinghua.edu.cn}

\author{Rae Ying Yee Wong}
\affiliation{\institution{Stanford University}\city{Stanford}\state{CA}\country{USA}}
\orcid{0009-0008-2677-0659}
\email{raewong@stanford.edu}

\author{Alan Zhu}
\affiliation{\institution{Carnegie Mellon University}\city{Pittsburgh}\state{PA}\country{USA}}
\orcid{0009-0001-8694-9246}
\email{aczhu@andrew.cmu.edu}

\author{Lijie Yang}
\affiliation{\institution{Carnegie Mellon University}\city{Pittsburgh}\state{PA}\country{USA}}
\orcid{0009-0004-7909-5416}
\email{lijiey@andrew.cmu.edu}

\author{Xiaoxiang Shi}
\affiliation{\institution{Shanghai Jiao Tong University}\city{Shanghai}\country{China}}
\orcid{0009-0000-6840-4691}
\email{lambda7shi@sjtu.edu.cn}

\author{Chunan Shi}
\affiliation{\institution{Peking University}\city{Beijing}\country{China}}
\orcid{0009-0009-7197-4965}
\email{spirited_away@pku.edu.cn}

\author{Zhuoming Chen}
\affiliation{\institution{Carnegie Mellon University}\city{Pittsburgh}\state{PA}\country{USA}}
\orcid{0009-0006-7797-573X}
\email{zhuominc@andrew.cmu.edu}

\author{Daiyaan Arfeen}
\affiliation{\institution{Carnegie Mellon University}\city{Pittsburgh}\state{PA}\country{USA}}
\orcid{0009-0009-5626-4551}
\email{marfeen@andrew.cmu.edu}

\author{Reyna Abhyankar}
\affiliation{\institution{University of California, San Diego}\city{San Diego}\state{CA}\country{USA}}
\orcid{0009-0005-6763-0108}
\email{vabhyank@ucsd.edu}

\author{Zhihao Jia}
\affiliation{\institution{Carnegie Mellon University}\city{Pittsburgh}\state{PA}\country{USA}}
\orcid{0000-0002-1270-5185}
\email{zhihao@cmu.edu}

\renewcommand{\shortauthors}{Miao$^*$, Oliaro$^*$, Zhang$^*$, Cheng$^*$ et al.}

\date{}

\begin{abstract}

This paper introduces \sys, a system that accelerates generative large language model (LLM) serving with \rev{{\em tree-based} speculative inference and verification}.
The key idea behind \Sys is leveraging small speculative models to predict the LLM’s outputs; the predictions are organized as a token tree, whose nodes each represent a candidate token sequence.
The correctness of all candidate token sequences represented by a token tree is verified against the LLM in parallel using a novel tree-based parallel decoding mechanism.
\Sys uses an LLM as a token tree verifier instead of an incremental decoder, which significantly reduces the end-to-end latency and computational requirement for serving generative LLMs while provably preserving model quality. 
Our evaluation shows that \sys outperforms existing LLM serving systems by 1.5-2.8$\times$ for distributed LLM inference and by 2.6-3.5$\times$ for offloading-based LLM inference, while preserving the same generative performance.
\Sys is publicly available at \url{https://github.com/flexflow/FlexFlow/}

\end{abstract}


\maketitle

\pagestyle{plain}

\section{Introduction}
\label{sec:intro}
Generative large language models (LLMs), such as ChatGPT~\cite{brown2020language} and GPT-4~\cite{openai2023gpt4}, have proven to be powerful in various application domains, including question answering, program synthesis, and task automation~\cite{zhang2019pretraining, liu2021makes}.
However, it is challenging to quickly and cheaply serve these LLMs due to their large volume of parameters, complex architectures, and high computational requirements.
For example, the largest GPT-3 architecture has 175 billion parameters, which requires more than eight NVIDIA 40GB A100 GPUs to store in half-precision floating points, and takes several seconds to serve a single inference request~\cite{brown2020language}.

\begin{figure*}
    \centering
    \subfloat[Incremental decoding.]{
    \includegraphics[scale=0.40]{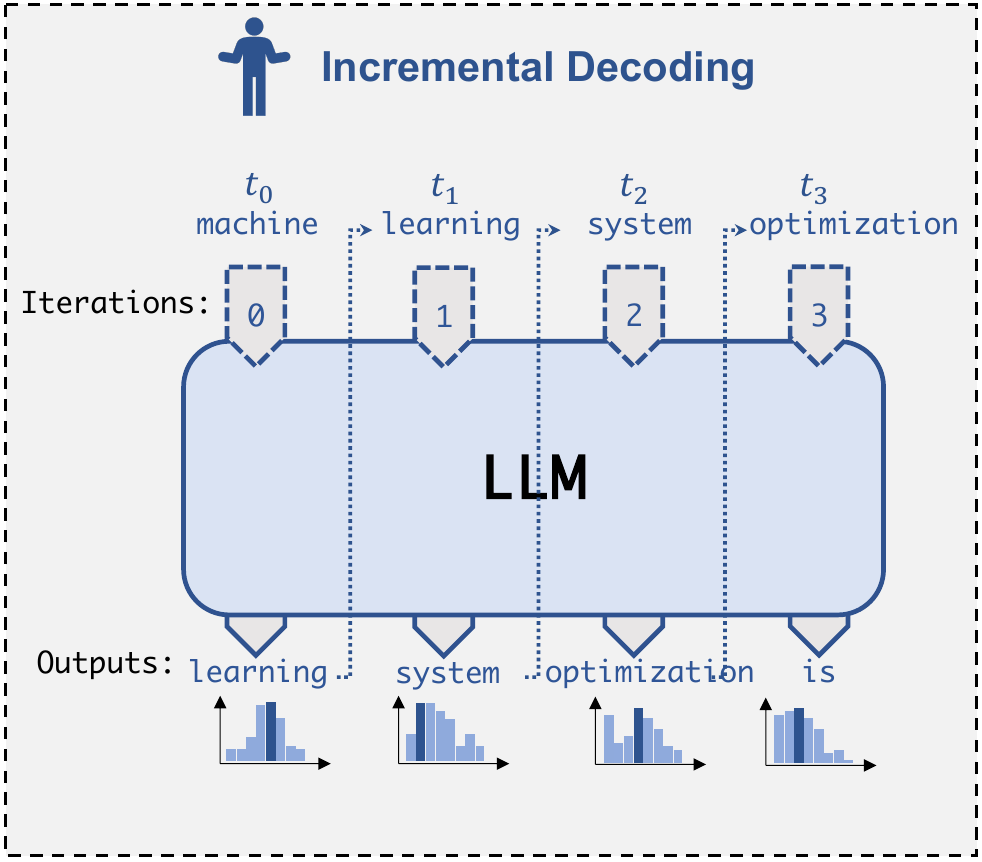}
    \label{fig:existing_approach}
    }
    \subfloat[\rev{Timeline Comparison.}]{
    \includegraphics[scale=0.40]{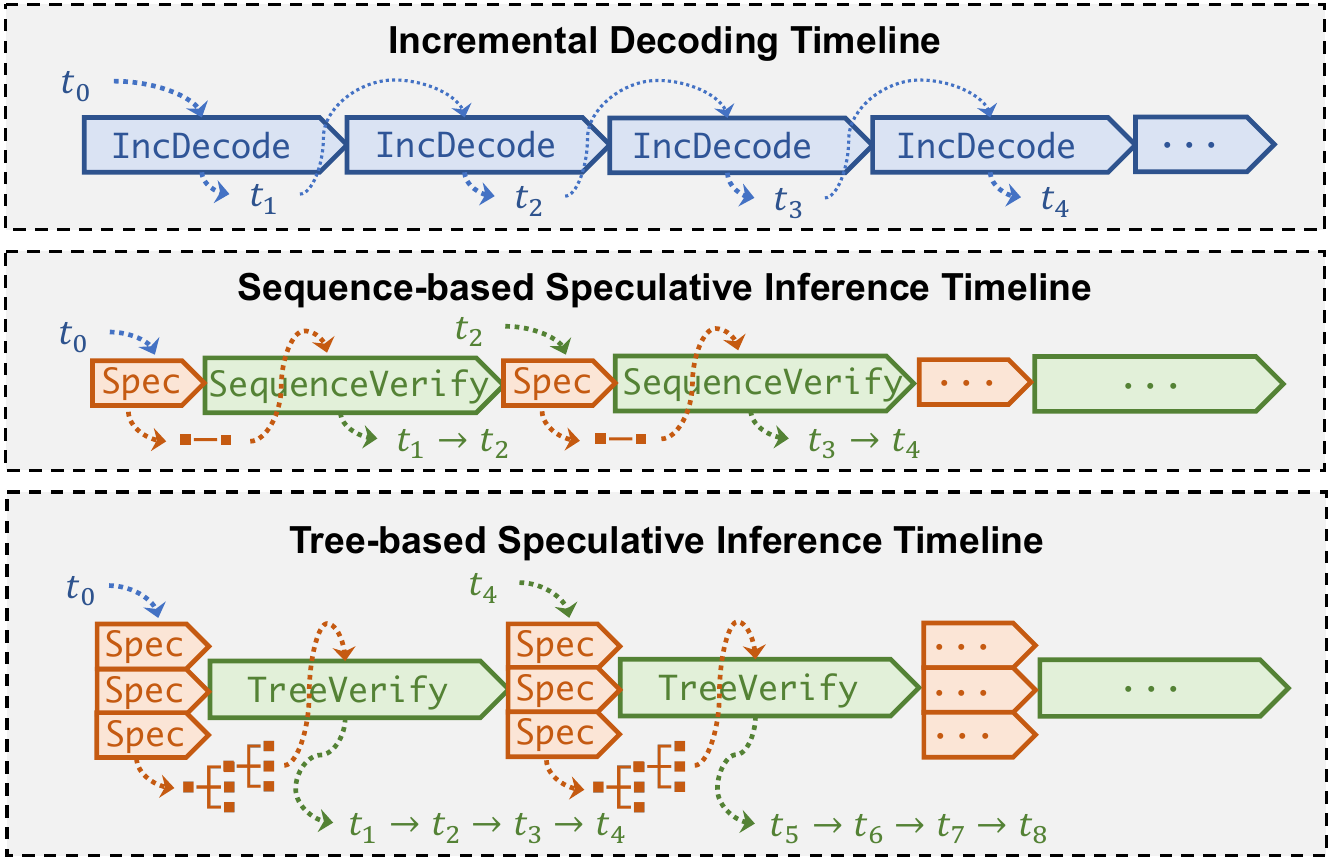}
    \label{fig:timeline}
    }
    \caption{\rev{Comparing the incremental decoding approach used by existing LLM serving systems, the sequence-based speculative inference approach, and the tree-based speculative inference approach used by \sys.}}
    \label{fig:specinfer_compared_to_incr_decoding}
\end{figure*}

An LLM generally takes as input a sequence of tokens, called {\em prompt}, and generates subsequent tokens one at a time, as shown in \Cref{fig:existing_approach}. 
The generation of each token in the sequence is conditioned on the input prompt and previously generated tokens and does not consider future tokens.
This approach is also called {\em autoregressive} decoding because each generated token is also used as input for generating future tokens.
This dependency between tokens is crucial for many NLP tasks that require preserving the order and context of the generated tokens, such as text completion~\cite{yu2022orca}.


Existing LLM systems generally use an {\em incremental decoding} approach to serving a request where the system computes the activations for all prompt tokens in a single step and then iteratively decodes {\em one} new token using the input prompt and all previously generated tokens~\cite{miao2023towards}. 
This approach respects data dependencies between tokens, but achieves suboptimal runtime performance and limited GPU utilization, since the degree of parallelism within each request is greatly limited in the incremental phase.
In addition, the attention mechanism of Transformer~\cite{vaswani2017attention} requires accessing the keys and values of all previous tokens to compute the attention output of a new token.
To avoid recomputing the keys and values for all preceding tokens, today's LLM systems use a caching mechanism to store their keys and values for reuse in future iterations.
For long-sequence generative tasks (e.g., GPT-4 supports up to 32K tokens in a request), caching keys and values introduces significant memory overhead, which prevents existing systems from serving a large number of requests in parallel due to the memory requirement of caching their keys and values.

\rev{Motivated by the idea of speculative execution in processor optimizations~\cite{gabbay1996speculative, smith1998study}, recent work introduces {\em sequence-based speculative inference}, which leverages a {\em small speculative model} (SSM) to generate a sequence of tokens and uses an LLM to examine their correctness in parallel~\cite{leviathan2022fast, xiaspeculative, stern2018blockwise, chen2023accelerating, gante2023assisted}.
These attempts only consider a token sequence generated by a single SSM for speculation, which cannot align well with an LLM due to the model capacity gap between them, since SSMs are generally orders of magnitude smaller than the LLM to maintain low memory and runtime overheads.
}
%



This paper introduces \sys, a system that improves the end-to-end latency and computational efficiency of LLM serving with {\em tree-based speculative inference and verification}.
\Cref{fig:timeline} illustrates a comparison between existing incremental decoding, sequence-based speculative inference, and our tree-based speculative inference.
A key insight behind \sys is to simultaneously consider a diversity of speculation candidates (instead of just one as in existing approaches) to maximize speculative performance.
%
These candidates are organized as a {\em token tree}, whose nodes each represents a sequence of speculated tokens.
The correctness of {\em all} candidate token sequences is verified against the LLM in parallel, which allows \sys to significantly increase the number of generated tokens in an LLM decoding step. 
\rev{Compared with sequence-based speculative inference, leveraging tree structures can significantly improve the success rate of verifying a token (e.g., from 52-57\% to 96-97\% for stochastic decoding as shown in \Cref{tab:acceptance-rate}).}
However, realizing this improvement requires addressing two unique challenges. Next, we elaborate on these challenges and the main ideas \sys uses to address them.

First, \sys must explore an extremely large search space of candidate token sequences to maximize speculative performance. While the idea of speculative execution has been widely deployed in a variety of optimization tasks in computer architecture and systems, including branch prediction in modern pipelined processors and value prediction for pre-fetching memory and files~\cite{gabbay1996speculative, smith1998study}, the search space considered by \sys is significantly larger due to two reasons:
(1) modern LLMs generally involve very large vocabularies, and (2) maximizing speculative performance requires predicting multiple future tokens (instead of just the next token). For example, all LLMs in the OPT model family consider 50,272 different possible tokens in their vocabulary, while \sys can correctly predict the next 4 tokens on average. Achieving this goal requires considering a search space of $50272^4\approx6\times 10^{18}$ different combinations of tokens. 

\sys leverages existing distilled, quantized, and/or pruned variants of an LLM, which we call small speculative models (SSMs), to guide speculation. 
A key challenging of using SSMs for speculative inference is that the alignment between an SSM and an LLM is inherently bounded by the model capacity gap, since an SSM is generally 100-1000$\times$ smaller than an LLM.
\rev{Instead of using a single SSM for sequence-based speculation, \sys maximizes speculative performance by simultaneously considering a variety of token sequences organized in a tree structure for a given input prompt.
\Sys introduces an {\em expansion}- and a {\em merge}-based mechanism for constructing token trees by exploiting diversity within a single SSM and across multiple SSMs, respectively.}
%

A second challenge \Sys must address is verifying the speculated tokens. Many LLM applications perform {\em stochastic decoding}, which samples the next token from a probability distribution instead of deterministically generating a token. 
To preserve an LLM's generative performance, \Sys must guarantee that its tree-based speculative inference and verification mechanism generates the next token by following the {\em exact same} probability distribution as incremental decoding.
To achieve this goal, we propose {\em multi-step speculative sampling}, a new sampling approach for SSMs that guarantees equivalence while maximizing the number of speculated tokens that can be verified.
To minimize the token tree verification cost, \sys introduces a {\em tree-based parallel decoding} mechanism, {\em simultaneously} verifying all tokens of a token tree against the LLM's output in a {\em single} LLM decoding step.

By leveraging tree-based speculative inference and verification, \Sys accelerates both distributed LLM inference across multiple GPUs and offloading-based LLM inference on one GPU.
Our evaluation shows that \Sys outperforms existing LLM serving systems by 1.5-2.8$\times$ for distributed LLM inference and by 2.6-3.5$\times$ for offloading-based LLM inference, while preserving the same generative accuracy.
%

To summarize, we make the following contributions:
\begin{itemize}
    \item We present \sys, a tree-based speculative inference and verification system for LLM serving.
    \item To maximize speculative performance, we propose a merge- and an expansion-based method to construct token trees by exploiting diversity within and across SSMs, respectively.
    \item To minimize verification cost, we introduce a tree-based parallel decoding mechanism to simultaneously verify all tokens of a token tree.
    \item We evaluate \sys and show that it outperforms existing systems by up to 2.8$\times$ for distributed inference and by up to 3.5$\times$ for offloading-based inference.
\end{itemize}

\section{\Sys's Overview}
\label{sec:overview}

\begin{figure*}[t]
    \centering
    \includegraphics[scale=0.4]{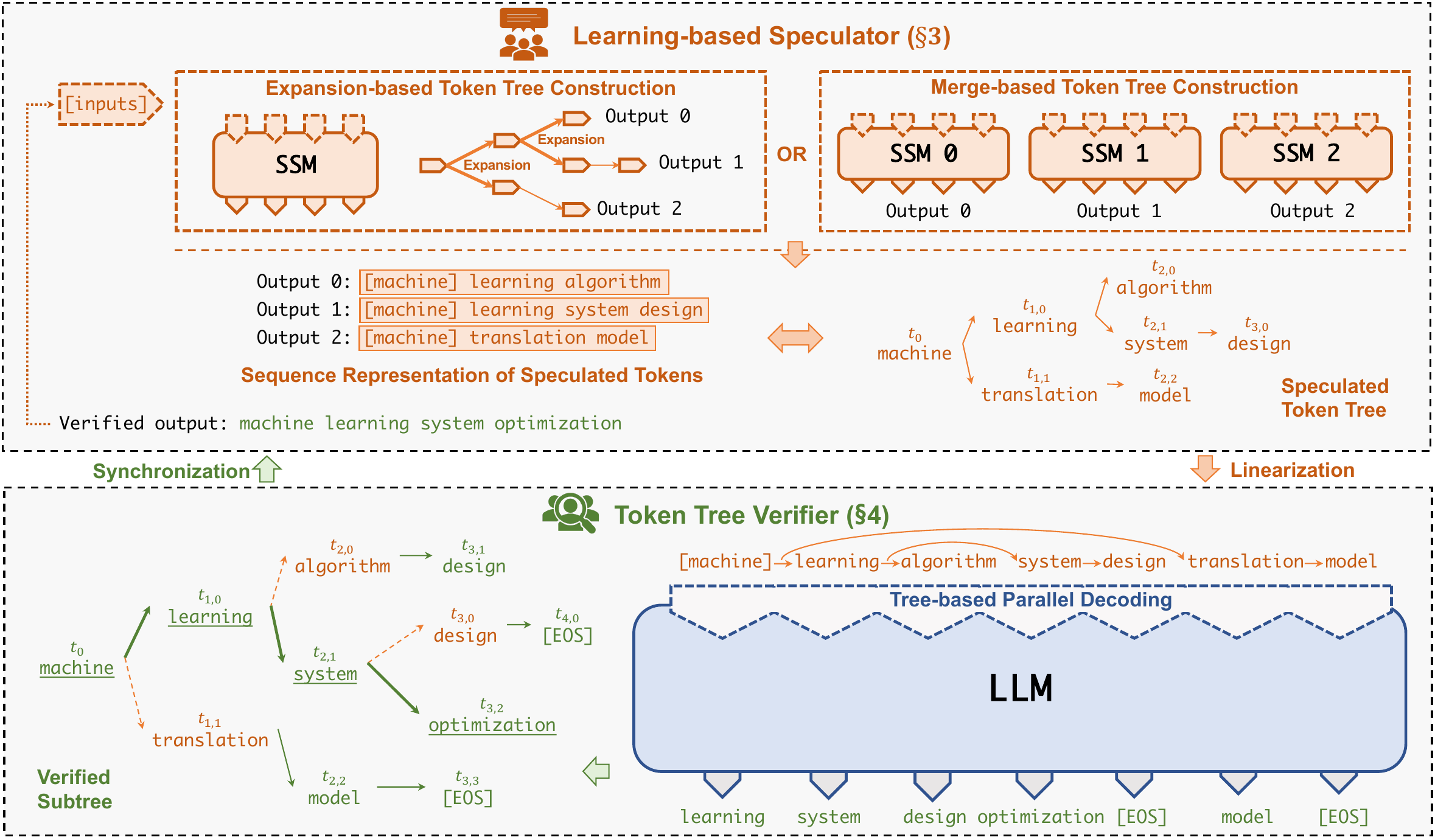}
    \caption{An overview of \Sys's tree-based speculative inference and verification mechanism.}
    \label{fig:speculative_inference}
\end{figure*}

\begin{algorithm}
\caption{The incremental decoding algorithm used in existing LLM serving systems.}
\label{alg1}
\begin{algorithmic}[1]
\State {\bf Input:} A sequence of input tokens $\m{I}$
\State {\bf Output:} A sequence of generated tokens
\State $\m{S} = \m{I}$
\While{true}
\State $t = \Call{Decode}{\textrm{LLM}, \m{S}}$
\State $\m{S}$.append($t$)
\If{$t = \langle \textrm{EOS} \rangle$}
\State {\bf Return} $\m{S}$
\EndIf
\EndWhile
\end{algorithmic}
\end{algorithm}

\begin{algorithm}
\caption{The speculation and verification algorithm used by \Sys. \textproc{Speculate} takes the current token sequence $\m{S}$ as an input and generates a speculated token tree $\m{N}$. 
\textproc{TreeParallelDecode} generates a token $\m{O}(u)$ for each node $u \in \m{N}$.
\textproc{VerifyGreedy} and \textproc{VerifyStochastic} examine $\m{N}$ against $\m{O}$ and produce a sequence of verified tokens $\m{V}$ using greedy or stochastic sampling, respectively.}
\label{alg2}
\begin{algorithmic}[1]
\State {\bf Input:} A sequence of input tokens $\m{I}$
\State {\bf Output:} A sequence of generated tokens
\State $\m{S} = \m{I}$
\While{true}
\State $\m{N} = \Call{Speculate}{\m{S}}$
\State $\m{O} = \Call{TreeParallelDecode}{\textrm{LLM}, \m{N}}$ 
\If{use greedy decoding}
\State $\m{V} = \Call{VerifyGreedy}{\m{O}, \m{N}}$
\Else
\State $\m{V} = \Call{VerifyStochastic}{\m{O}, \m{N}}$
\EndIf
\For{$t \in \m{V}$}
\State $\m{S}$.append($t$)
\If{$t = \langle \textrm{EOS} \rangle$}
\State {\bf return} $\m{S}$
\EndIf
\EndFor
\EndWhile
\State

\Function{VerifyGreedy}{$\m{O}, \m{N}$}
\State $\m{V} = \emptyset$, $u \leftarrow$ the root of token tree $\m{N}$
\While{$\exists v \in \m{N}. p_v = u \textrm{ and } t_v = \m{O}(u)$}
\State $\m{V}$.append($t_v$) 
\State $u = v$ 
\EndWhile
\State $\m{V}$.append($\m{O}(u)$)
\State {\bf return} $\m{V}$
\EndFunction
\State

\Function{VerifyStochastic}{$\m{O}, \m{N}$}
\State $\m{V} = \emptyset$, $u \leftarrow$ the root of token tree $\m{N}$
\While{$u$ is a non-leaf node}
\State $\m{H} = \textrm{child}(u)$
\Comment{\textcolor{violet}{The set of child nodes for $u$}}
    \While{$\m{H}$ is not empty}
    \State $s\sim \textrm{rand}(\m{H}), r \sim U(0, 1), x_s=\m{H}[s]$ 
        \If{$r \leq {P(x_s \mid u, \Theta_{LLM})} / {P(x_s \mid u, \Theta_{SSM_s})}$}
            \LineComment{\textcolor{violet}{Token $x_s$ passes verification.}}
            \State $\m{V}$.append($x_s$)
            \State $u = s$
            \State {\bf break}
        \Else            
            \LineComment{\textcolor{violet}{Normalize the residual $P(x \mid u, \Theta_{LLM})$}}
            \State $P(x \mid u, \Theta_{\textrm{LLM}})\coloneqq \textrm{norm}(\max(0, P(x \mid u, \Theta_{\textrm{LLM}})-P(x \mid u, \Theta_{\textrm{SSM}_s})))$ 
            \State $\m{H}.\textrm{pop}(s)$
        \EndIf   
    \EndWhile
\If{$\m{H}$ is empty}
    \State {\bf break}
\EndIf
\EndWhile
\LineComment{All SSMs fail verification; sample the next token}
\State $x_{\textrm{next}} \sim P(x \mid u, \Theta_{LLM})$
\State $\m{V}$.append($x_{\textrm{next}}$)
\State {\bf return} $\m{V}$

\EndFunction
\end{algorithmic}
\end{algorithm}

\Cref{fig:speculative_inference} shows an overview of \sys, which
includes a {\em learning-based speculator} that takes as input a sequence of tokens, and produces a {\em speculated token tree}.
The goal of the speculator is to predict the LLM's output by maximizing the overlap between the speculated token tree and the tokens generated by the LLM using incremental decoding (Alg.~\ref{alg1}).

There are several ways to prepare SSMs for speculative inference. 
First, modern LLMs generally have many smaller architectures pre-trained together with the LLM using the same datasets. 
For example, in addition to the OPT-175B model with 175 billion parameters, the OPT model family also includes OPT-125M and OPT-350M, two variants with 125 million and 350 million parameters, which were pre-trained using the same datasets as OPT-175B~\cite{zhang2022opt}.
These pre-trained small models can be directly used as SSMs. 
\rev{Second, to improve the coverage of speculated tokens from SSMs, \Sys takes an expansion-based and a merge-based speculation method as shown at the top of \Cref{fig:speculative_inference}. The speculated tokens are organized in a token tree structure.}

\Sys's usage of an LLM is also different from that of existing LLM serving systems. 
Instead of using the LLM as an incremental decoder that predicts the next single token, \sys uses the LLM as a token tree verifier that verifies a speculated token tree against the LLM's output.
For each token, \sys computes its activations by considering all of its ancestors in the token tree as its preceding tokens.
For example, in \Cref{fig:speculative_inference}, the attention output of the token $t_{3,0}$ is calculated based on sequence $(t_0, t_{1,0}, t_{2, 1}, t_{3,0})$, where $t_0$, $t_{1,0}$, and $t_{2,1}$ are $t_{3,0}$'s ancestors in the token tree.
\Sys includes a novel tree-based parallel decoding mechanism to simultaneously verify {\em all} tokens of a token tree in a single LLM decoding step.

\Sys's speculative inference and token tree verification provide two key advantages over the incremental decoding approach of existing LLM inference systems.

\paragraph{Reduced memory accesses to LLM parameters.} The performance of  LLM inference is largely limited by accesses to GPU memory.
In the existing incremental decoding approach, generating a single token requires accessing all parameters of an LLM. 
The problem is exacerbated for offloading-based LLM inference systems, which use limited computational resources such as a single commodity GPU to serve LLMs by utilizing CPU DRAM and persistent storage to save model parameters and loading these parameters to GPU's high bandwidth memory (HBM) for computation.
Compared to the incremental decoding approach, \sys significantly reduces accesses to LLM parameters whenever the overlap between a speculated token tree and the LLM's actual output is not empty.
Reduced accesses to GPU device memory and reduced data transfers between GPU and CPU memory can also directly translate to decreased energy consumption, since accessing GPU HBM consumes two or three orders of magnitude more energy than floating point arithmetic operations.

\paragraph{Reduced end-to-end inference latency.}
Serving LLMs suffers from long end-to-end inference latency.
For example, the GPT-3 architecture includes 175 billion parameters and requires many seconds to serve a request.
In the existing incremental decoding approach, the computation for generating each token depends on the keys and values of all previously generated tokens, which introduces sequential dependencies between tokens and requires modern LLM serving systems to serialize the generation of different tokens for each request.
In \sys, LLMs are used as a verifier that takes a speculated token tree as an input and can simultaneously examine {\em all} tokens in the token tree by making a single verification pass over the LLM. 
This approach enables parallelization across different tokens in a single request and reduces the LLM's end-to-end inference latency.


\section{Learning-based Speculator}
\label{sec:speculation}

\begin{table}
\begin{center}
\caption{\rev{The success rate of verifying a token for \llama-7B using the top-$k$ tokens derived from \llama-68M. The five prompt datasets are described in \Cref{subsec:setup}.}}
\label{tab:acceptance-rate}
\begin{tabular}{l|l|ccccc}
\toprule
& Dataset     & $k=1$ & $k=2$ & $k=3$ & $k=4$ & $k=5$ \\ \midrule
\multirow{5}{*}{\rotatebox[origin=c]{90}{\parbox{2cm}{\bf \centering Greedy\\ decoding}}}
& Alpaca     &68\%          & 77\%    & 81\%     & 84\%     & 85\%     \\
& CP         &69\%          & 79\%    & 83\%     & 86\%     & 87\%     \\
& WebQA      &62\%          & 72\%    & 77\%     & 80\%     & 82\%     \\ 
& CIP        &70\%          & 81\%    & 85\%     & 88\%     & 89\%     \\
& PIQA       &63\%          & 75\%    & 79\%     & 83\%     & 85\%     \\
\midrule
\multirow{5}{*}{\rotatebox[origin=c]{90}{\parbox{2cm}{\bf \centering Stochastic\\ decoding}}}
& Alpaca     &54\%          & 81\%    & 91\%     & 95\%     & 97\%     \\
& CP         &56\%          & 82\%    & 92\%     & 95\%     & 97\%     \\
& WebQA      &52\%          & 80\%    & 90\%     & 94\%     & 96\%     \\ 
& CIP        &57\%          & 84\%    & 92\%     & 95\%     & 97\%     \\
& PIQA       &55\%          & 82\%    & 91\%     & 95\%     & 97\%     \\
\bottomrule
\end{tabular}
\end{center}
\end{table}

\rev{Existing speculative decoding methods perform {\em sequence-based} speculation, where an SSM predicts a single sequence of tokens to be verified by an LLM.
However, a key limitation of a single speculated sequence is that the probability of a successful alignment between the LLM and the speculated token sequence decays exponentially with the expected alignment length.
This can be further exacerbated by the fact that the speculation only includes a single candidate token to verify per step, resulting in suboptimal speculative performance.
On the other hand, by encouraging more diverse speculated candidates per step, the probability of a successful match per step (i.e., the token decoded by the LLM is in this candidate pool) can be greatly improved. 
To this end, \sys aims to construct a tree of speculated candidates by exploiting diversity within a single SSM and across multiple SSMs.
In particular, \sys's {\em learning-based speculator} aggregates the predictions of one or multiple SSMs to maximize speculative performance while maintaining low memory overhead and inference latency.
\Sys uses a {\em token tree} to organize the tokens produced by the speculator and introduces two methods for constructing token trees: {\em expansion-} and {\em merge-based} tree constructions.}

\begin{definition}[Token Tree]
A token tree $\m{N}$ is a tree structure, where each node $u \in \m{N}$ is labeled with a token $t_u$, and $p_u$ represents $u$'s parent node in the token tree. 
For each node $u$, $S_u$ represents a sequence of tokens identified by concatenating $S_{p_u}$ and $\{t_u\}$\footnote{For the root node $r$, $S_r$ represents the token sequence $\{t_r\}$.}.
\end{definition}

\rev{\paragraph{Expansion-based token tree construction.}
One approach to creating a token tree involves deriving {\em multiple} tokens from an SSM within a single decoding step.
This approach is motivated by an important observation that when an SSM misaligns with an LLM (i.e., the two models select different top-1 tokens), the token selected by the LLM is generally among the top-$k$ tokens from the SSM for very small values of $k$.
\Cref{tab:acceptance-rate} shows the success rate of verifying a token using the top-$k$ tokens derived from an SSM, where a verification is successful if the token selected by the LLM is among the top-$k$ tokens from the SSM. Compared to only using the top-$1$ token from an SSM, using the top-$5$ tokens can increase the success rate from 70\% to 89\% for greedy decoding and from 57\% to 97\% for stochastic decoding.}

\begin{figure}
    \centering
    \includegraphics[width=\linewidth]{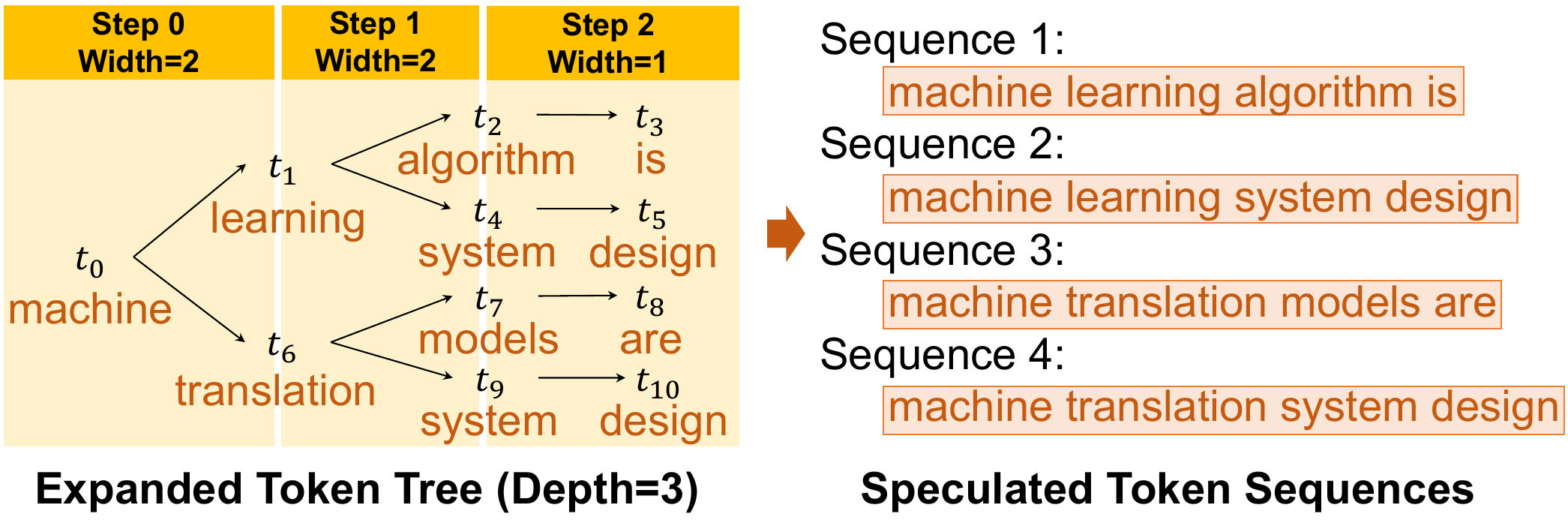}
    \caption{\rev{Illustration of token tree expansion.}}
    \label{fig:token_tree_merge}
\end{figure}

\rev{Directly selecting the top-$k$ tokens at each step leads to an exponential increase in the number of potential token sequences, which substantially elevates inference latency and memory overhead. Consequently, we adopt a {\em static} strategy that expands the token tree following a preset {\em expansion configuration} represented as a vector of integers $\langle k_1, k_2, ..., k_m\rangle$, where $m$ denotes the maximum number of speculative decoding steps, and $k_i$ indicates the number of tokens to expand for each token in the $i$-th step.
For example, \Cref{fig:token_tree_merge} illustrates the expansion configuration $\langle 2, 2, 1 \rangle$, leading to four token sequences.
Our evaluation (see \Cref{subsec:eval_token_tree}) shows that even a simple strategy can generate highly accurate speculative results.
We acknowledge that {\em dynamically} expanding a token tree from an SSM is an opening research problem beyond the scope of this paper, which we leave as future work.}


\paragraph{Merge-based token tree construction.}
In addition to using a single SSM, \sys can also combine multiple SSMs to jointly predict an LLM's output. \Sys uses an unsupervised method to {\em collectively boost-tune} a pool of SSMs to align their outputs with that of the LLM by leveraging adaptive boosting~\cite{freund1999short}.
\Sys uses SSMs to predict the next few tokens that an LLM will generate, and uses general text datasets (e.g., the OpenWebText corpus~\cite{Gokaslan2019OpenWeb} in our evaluation) to adaptively align the aggregated output of multiple SSMs with the LLM in a fully unsupervised fashion.
In particular, \Sys converts a text corpus into a collection of prompt samples and use the LLM to generate a token sequence for each prompt. 
\Sys first fine-tunes one SSM at a time to the fullest and marks all prompt samples where the SSM and LLM generate identical subsequent tokens.
Next, \Sys filters all marked prompt samples and uses all remaining samples in the corpus to fine-tune the next SSM to the fullest. 

By repeating this process for every SSM in the pool, \Sys obtains a diverse set of SSMs whose aggregated output largely overlaps with the LLM's output on the training corpus.
All SSMs have identical inference latency, and therefore running all SSMs on different GPUs in parallel does not increase the latency of speculative inference compared to using a single SSM. 
In addition, \Sys uses data parallelism to serve SSMs across multiple GPUs, and therefore using multiple SSMs does not increase the memory overhead on each GPU.
In the case where multiple SSMs are employed, the output of each SSM is considered as a token tree, and \Sys performs {\em token tree merge} to aggregate all speculated tokens in a single tree structure.

\begin{definition}[Token Tree Merge]
$\m{M}$ is the tree merge of $m$ token trees $\{\m{N}_i\}$ ($1\leq i \leq m$) if and only if $\forall 1\leq i \leq m, \forall u \in \m{N}_i, \exists v \in \m{M}$ such that $S_v = S_u$ and vice versa.
\end{definition}

Intuitively, each token tree represents a set of token sequences. Merging multiple token trees produces a new tree that includes all token sequences of the original trees.
For example, \Cref{fig:token_tree_merge} shows the token tree derived by merging four sequences of tokens. Each token sequence is identified by a node in the merged token tree.

\rev{Note that, in addition to boosting, there are several other ensemble learning methods (e.g., voting, bagging, and stacking)~\cite{ganaie2022ensemble} that can be used to combine the outputs from multiple SSMs, and we leave the exploration as future work.}

\section{Token Tree Verifier}
\label{sec:verification}

This section introduces \sys's {\em token tree verifier}, which takes as input a token tree generated by the speculator and verifies the correctness of its tokens against an LLM's output.
A key idea behind the design of \Sys is {\em simultaneously} verifying all sequences of a token tree against the original LLM's output by making a {\em single} pass over the LLM's parameters.
This functionality allows \Sys to opportunistically decode {\em multiple} tokens (instead of a single token in incremental decoding), resulting in reduced memory accesses to the LLM's parameters.
A challenge \Sys must address in token tree verification is efficiently computing the attention scores for {\em all} sequences of a token tree.
To this end, we introduce {\em tree attention}, which generalizes the attention mechanism~\cite{vaswani2017attention} from sequence to tree structure.
In addition, we develop a {\em tree-based parallel decoding} mechanism that can decode {\em all} tokens in a token tree in parallel.

\S\ref{subsec:tree_attention} and \S\ref{subsec:optimization} describe tree attention and tree-based parallel decoding. 
\S\ref{subsec:verification} introduces the mechanism to verify a token tree against the LLM's output.

\subsection{Tree Attention}
\label{subsec:tree_attention}
Transformer-based language models use the attention mechanism to reason about sequential information~\cite{vaswani2017attention}.
LLMs generally use decoder-only, multi-head self-attention, which takes a single input tensor $X$ and computes an output tensor $O$ via scaled multiplicative formulations as follows.
\begin{eqnarray}
    Q_i = X \times W_i^Q, & K_i = X \times W_i^K, \\
    V_i = X \times W_i^V, & A_i = \frac{(Q_i \times K_i^T)}{\sqrt{d}}, \\
    H_i = \textrm{softmax}\big(\textrm{mask}(A_i)\big) V_i, & O = (H_1, ...,H_h) W^O 
    \label{eqn2}
\end{eqnarray}
where $Q_i$, $K_i$, and $V_i$ denote the query, key, and value tensors of the $i$-th attention head ($1\leq i \leq h$), $W_i^Q$, $W_i^K$, and $W_i^V$ are the corresponding weight matrices. 
$A_i$ is an $l\times l$ matrix that represents the attention scores between different tokens in the input sequence, where $l$ is the sequence length. To preserve causality when generating tokens (i.e., a token in the sequence should not affect the hidden states of any preceding tokens), the following causal mask function is applied:
\begin{equation}
    \textrm{mask}(A)_{jk} =     \begin{cases}
    A_{jk} & j \geq k\\
    - \infty & j < k 
    \end{cases}.
\end{equation}
Intuitively, when computing the attention output of the $j$-th token in the sequence, all subsequent tokens should have an attention score of $-\infty$ to indicate that the subsequent tokens will not affect the attention output of the $j$-th token\footnote{Note that we use $-\infty$ (instead of 0) to guarantee that the softmax's output is 0 for these positions.}. 
In Equation~\ref{eqn2}, $H_i$ represents the output of the $i$-th attention head, and $W_O$ is a weight matrix used for computing the final output of the attention layer.

Note that the attention mechanism described above applies only to a sequence of tokens. We generalize the attention mechanism to arbitrary tree structures.

\begin{definition}[Tree Attention]
For a token tree $\m{N}$ and an arbitrary node $u \in \m{N}$, its tree attention is defined as the output of computing the original Transformer-based sequence attention on $S_u$ (i.e., the token sequence represented by $u$):
\begin{equation}
    \textproc{TreeAttention}(u) = \textproc{Attention}(S_u) \forall u \in \m{N}
\end{equation}
\end{definition}
For a given set of token sequences, since each sequence $S$ is covered by a node of the merged token tree, performing tree attention on the token tree allows \Sys to obtain the attention output for {\em all} token sequences. 

\subsection{Tree-based Parallel Decoding}
\label{subsec:optimization}

\begin{figure*}
    \centering
    \includegraphics[width=0.9\textwidth]{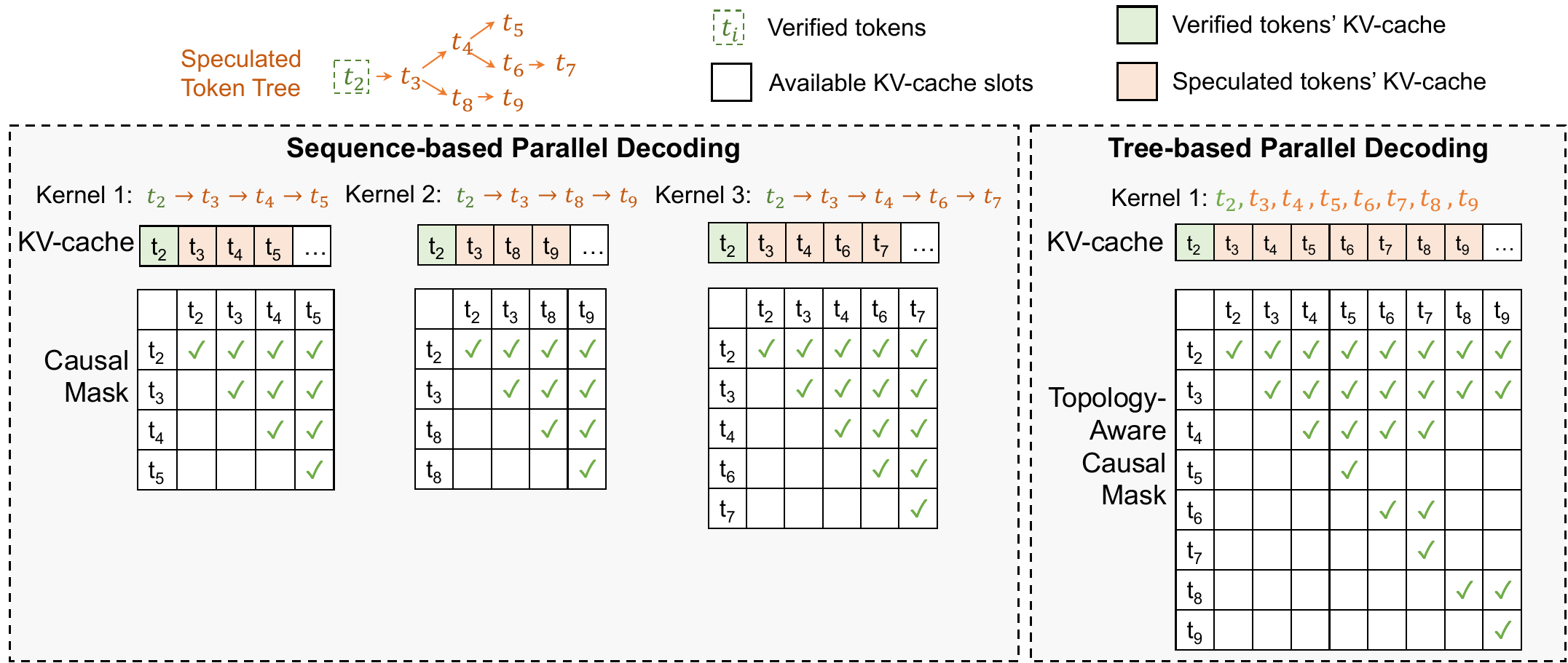}
    \caption{Comparing \Sys's tree-based parallel decoding with existing sequence-based decoding.}
    \label{fig:kv_cache}
\end{figure*}

This section describes \sys's {\em tree-based parallel decoding} mechanism for computing tree attention for {\em all} tokens in a token tree {\em in parallel}.
A key challenge \Sys must address in computing tree attention is managing {\em key-value cache}. 
In particular, the attention mechanism of Transformer~\cite{vaswani2017attention} requires accessing the keys and values of all preceding tokens to compute the attention output of each new token, as shown in Equation~\ref{eqn2}.
To avoid recomputing these keys and values, today's LLM inference systems generally cache the keys and values of all tokens for reuse in future iterations, since the causal relation guarantees that a token's key and value remain unchanged in subsequent iterations (i.e., $\rm{mask(A)_{jk}} = -\infty$ for any $j < k$).
However, when computing tree attention, different sequences in a token tree may include conflicting key-value caches. For example, for the speculated token tree in \Cref{fig:kv_cache}, two token sequences $(t_2, t_3, t_4, t_5)$ and $(t_2, t_3, t_8, t_9)$ have different keys and values for the third and fourth positions.

A straightforward approach to supporting key-value cache is employing the sequence-based decoding of existing LLM inference systems and using a different key-value cache for each sequence of a token tree, as shown on the left of \Cref{fig:kv_cache}.
However, this approach is computationally very expensive and involves redundant computation, since two token sequences sharing a common prefix have the same attention outputs for the common prefix due to the causal mask in Equation~\ref{eqn2}.
In addition, launching one kernel for each token sequence introduces additional kernel launch overhead.

\Sys introduces two key techniques to realize tree-based parallel decoding.

\paragraph{Depth-first search to update key-value cache.} Instead of caching the keys and values for individual token sequences of a token tree, \Sys reuses the same key-value cache across all token sequences by leveraging a {\em depth-first search} mechanism to traverse the token tree, as shown in \Cref{fig:kv_cache}, where \Sys visits $t_2,t_3,...,t_9$ by following a depth-first order to traverse the token tree and update the shared key-value cache.
This approach allows \Sys to maintain the correct keys and values for all preceding tokens when computing the attention output of a new token.

\paragraph{Topology-aware causal mask.} A straightforward approach to computing tree attention is calculating the tree attention output for individual tokens by following the depth-first order described earlier. 
However, this approach would result in high GPU kernel launch overhead since each kernel only computes tree attention for one token sequence.
In addition, executing these kernels in parallel requires additional GPU memory to store their key-value caches separately due to cache conflict.
A key challenge that prevents \Sys from batching multiple tokens is that the attention computation for different tokens requires different key-value caches and therefore cannot be processed in parallel.

We introduce {\em topology-aware casual mask} to fuse tree attention computation of all tokens in a single kernel. 
To batch attention computation, \Sys uses a tree topology instead of the original sequence topology to store the keys and values of all tokens in a token tree in the key-value cache.
For example, to compute tree attention for the speculated token tree shown in \Cref{fig:kv_cache}, \Sys takes both verified tokens (i.e., $t_2$) and all speculated tokens (i.e., $t_3,t_4,...,t_9$) as inputs.
This approach allows \Sys to fuse the attention computation into a single kernel but also results in attention scores that violate the causal dependency (e.g., $t_7$'s attention computation uses all previous tokens, including $t_5$ which is not in $t_7$'s token sequence).
To fix the attention scores for these pairs, \Sys updates the causal mask based on the token tree's topology.
This approach computes the exact same attention output as incremental decoding, while resulting in much fewer kernel launches compared to sequence-based decoding.

\subsection{Token Verification}
\label{subsec:verification}

For a given speculated token tree $\m{N}$, 
\Sys uses tree-based parallel decoding (see \Cref{subsec:optimization}) to compute its tree attention and generate an output tensor $\m{O}$ that includes a token for each node $u\in\m{N}$.
Next, \Sys's {\em token tree verifier} examines the correctness of speculated tokens against the LLM. 
\Sys supports both greedy and stochastic sampling as shown in~\Cref{alg2}. 
\begin{figure}
    \centering
    \includegraphics[scale=0.32]{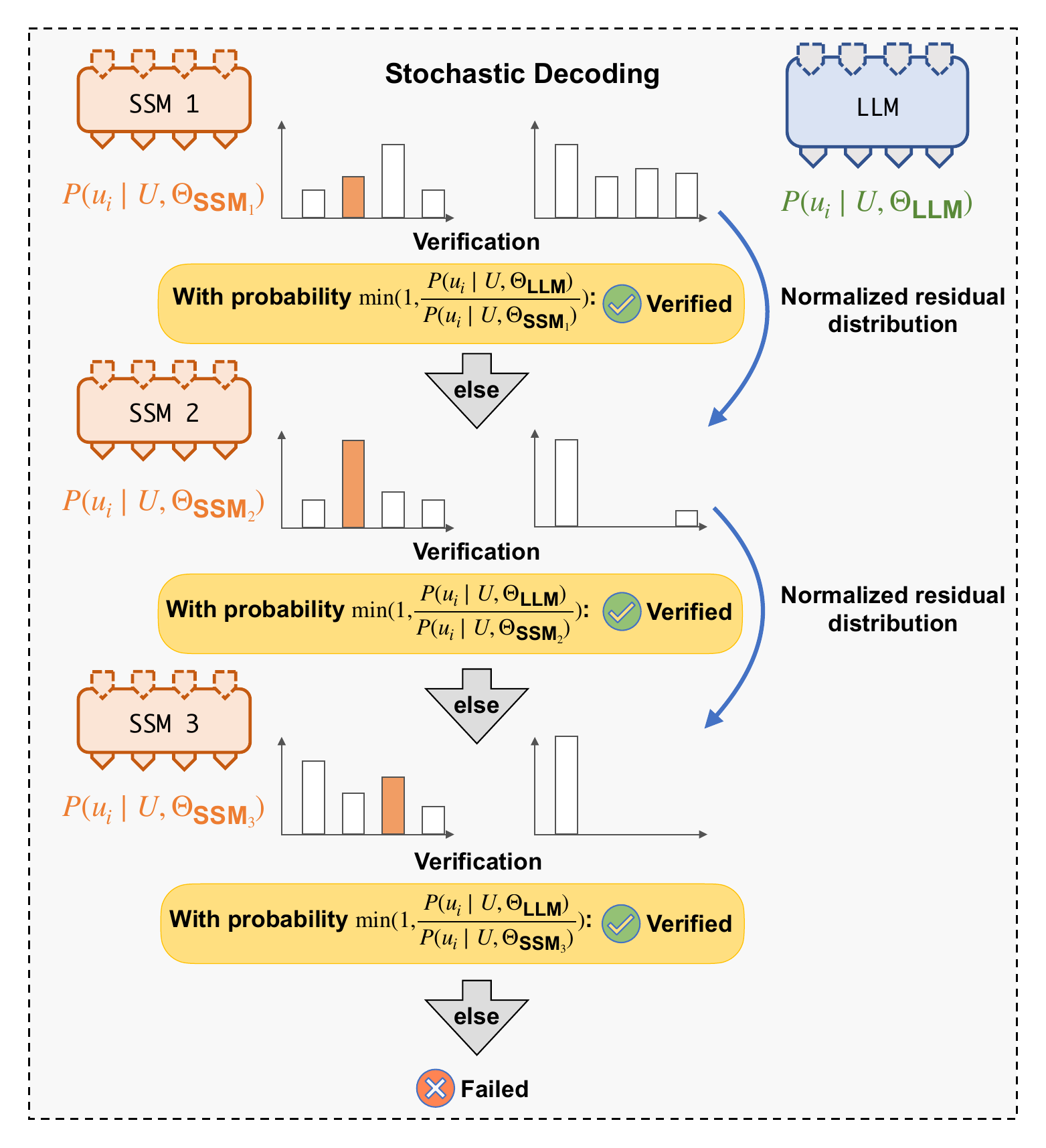}
    \caption{Illustrating the multi-step speculative sampling mechanism for verifying LLMs with stochastic sampling.}
    \label{fig:decodings}
\end{figure}

\paragraph{Greedy decoding.}
Many LLM applications generate tokens using {\em greedy decoding}, which greedily selects the token with the highest likelihood in each decoding step.
The $\textproc{VerifyGreedy}$ function in \Cref{alg2} shows how \Sys verifies a speculated token tree $\m{N}$ with greedy decoding.
\Sys starts from the root of $\m{N}$ and iteratively examines a node's speculated results against the LLM's original output.
For a node $u\in\m{N}$, \Sys successfully speculates its next token if $u$ includes a child node $v$ (i.e., $p_v = u$) whose token matches the LLM's output (i.e., $t_v = \m{O}(u)$).
In this case, \Sys finishes its verification for node $u$ and moves on to examine its child $v$.
When the node $u$ does not include a child that contains the LLM's output, \Sys adds $\m{O}(u)$ as a verified node in $\m{N}$ and terminates the verification process.
Finally, all verified nodes are appended to the current generated token sequence $\m{V}$.
Token tree verification allows \Sys to opportunistically decode multiple tokens (instead of a single token in the incremental decoding approach), while preserving the same generative performance as incremental decoding.

\paragraph{Stochastic decoding.}
To improve the diversity of generated tokens, many LLM applications perform {\em stochastic decoding}, which samples a token from a probability distribution $P(u_i | u_0, ..., u_{i-1}; \Theta_{LLM})$, where $U = u_0,...,u_{i-1}$ are previously generated tokens, $u_i$ is the next token to generate, and $\Theta_{LLM}$ represents a parameterized LLM.

To verify a speculated token tree with stochastic decoding, we introduce a {\em multi-step speculative sampling} (MSS) algorithm to conduct verification, whose pseudocode code is shown in the $\textproc{VerifyStochastic}$ function in \Cref{alg2} and illustrated in \Cref{fig:decodings}.
Our method provably preserves an LLM's generative performance as incremental decoding while optimizing the number of speculated tokens that can be verified.
\Cref{thm:stochastic_sampling} proves its correctness.

\begin{theorem}
\label{thm:stochastic_sampling}
For a given $LLM$ and $m$ SSMs (i.e., $SSM_1$,...,$SSM_m$, let $P(u_i | U; \Theta_{LLM})$ be the probability distribution of sampling a token using stochastic decoding, where $U = u_0,...,u_{i-1}$ are previously generated tokens, $u_i$ is the next token to generate, $\Theta_{LLM}$ represents the parameterized LLM. 

Let $P_{\rm{\Sys}}(u_i | U; \Theta_{LLM}, \{\Theta_{SSM_j}\})$ be the probability distribution of sampling token $u_i$ using \Sys's multi-step speculative sampling (see the $\textproc{VerifyStochastic}$ function in \Cref{alg2}), where $\Theta_{SSM_j}$ is the $j$-th parameterized SSM. Then $ \forall U, u_i, \Theta_{LLM}, \Theta_{SSM_j}$ we have
\begin{eqnarray}
P(u_i \mid U; \Theta_{LLM}) = P_{\rm{\Sys}}(u_i \mid U; \Theta_{LLM}, \{\Theta_{SSM_j}\})
\end{eqnarray}
\end{theorem}

A proof of this theorem is presented in~\cite{miao2023specinfer}.

We acknowledge that a more straightforward approach to preserving the probability distribution of stochastic decoding is directly sampling the next token $x \sim P(u_i \mid U; \Theta_{LLM})$ and examining whether $x$ is a child node of $u_{i-1}$ in the speculated token tree. We call this approach {\em naive sampling} (NS) and show that \Sys's multi-step speculative sampling has a uniformly lower rejection probability than naive sampling.

\begin{theorem}
\label{theorem:MSS}
    Let $P\left(\textrm{reject} \mid \textrm{MSS}, U, \Theta_{\textrm{LLM}}, \{\Theta_{\textrm{SSM}_j}\} \right)$ denote the probability of rejecting speculation following multi-step speculative sampling with abbreviation $P(\textrm{reject}\mid \textrm{MSS})$, and $P\left(\textrm{reject} \mid \textrm{NS}, U, \Theta_{\textrm{LLM}}, \{\Theta_{\textrm{SSM}_j}\}\right)$ the probability of rejecting speculation following Naive Sampling (NS) with abbreviation $P(\textrm{reject}\mid\textrm{NS})$. Then $\forall U, \Theta_{\textrm{LLM}}, \{\Theta_{\textrm{SSM}_j}\}$, we have $$P(\textrm{reject}\mid \textrm{MSS}) \leq P(\textrm{reject}\mid\textrm{NS})$$
\end{theorem}

We present a proof of~\Cref{theorem:MSS} in~\cite{miao2023specinfer}.

\rev{Note that prior work has introduced single-step speculative sampling for sequence-based speculative inference~\cite{leviathan2022fast, chen2023accelerating}. Different from these approaches, \Sys leverages token trees for improving speculative performance, which requires a different verification algorithm. As a result, \Sys performs multi-step verification (see \textproc{VerifyStochastic} in \Cref{alg2}) across all branches of a token to maximize the success rate while preserving equivalence as incremental decoding.
The proposed MSS algorithm not only works for  merge-based method with multiple SSMs, but also supports expansion-based method with one SSM and top-$k$ sampling.
}

\section{System Design and Implementation}

\begin{figure}
    \centering
    \includegraphics[scale=0.28]{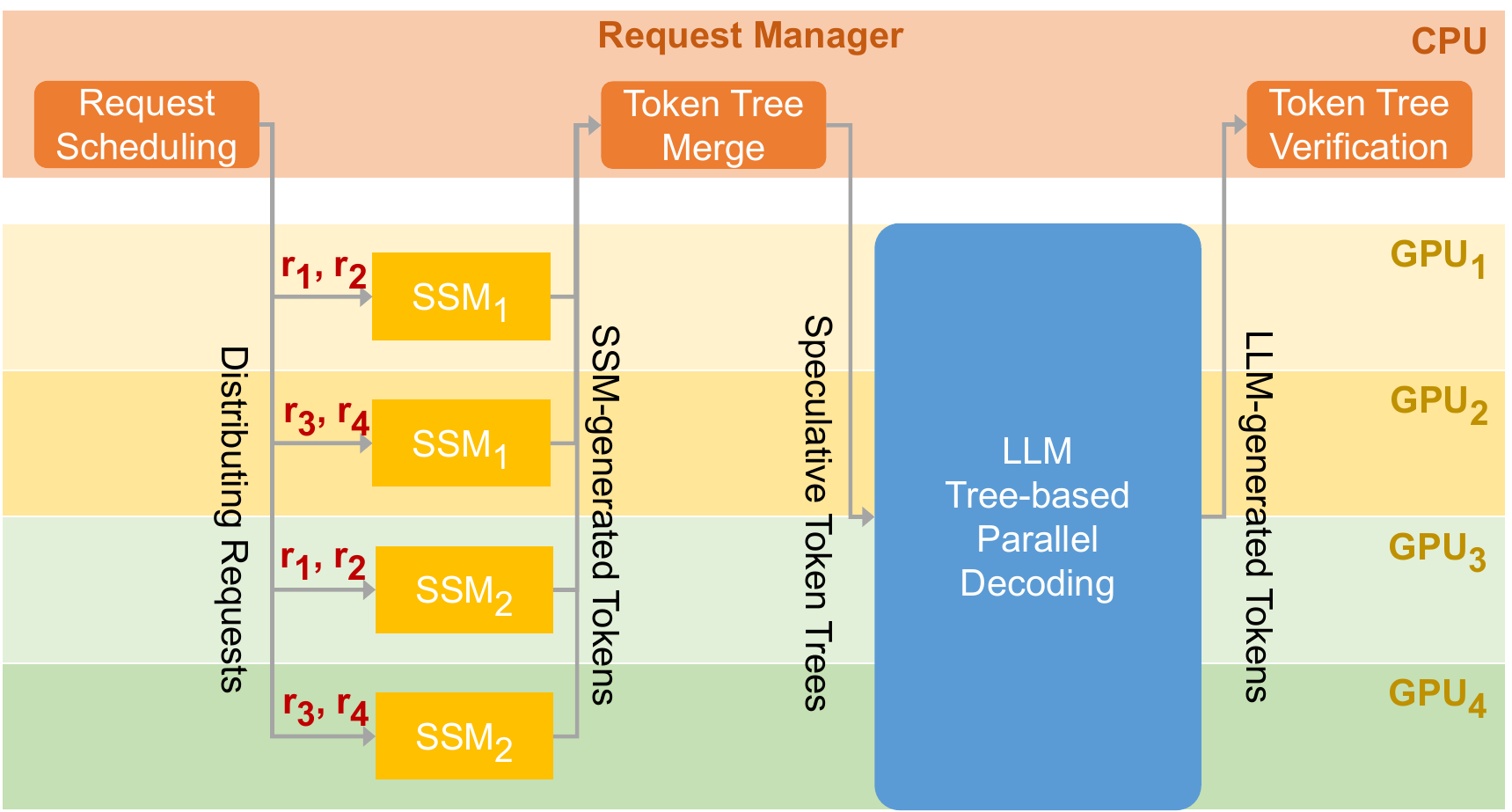}
    \caption{\Sys's workflow for one iteration of speculative inference and verification. \Sys uses data parallelism to serve SSMs, and combine tensor model parallelism and pipeline model parallelism for serving an LLM.}
    \label{fig:workflow}
\end{figure}

This section describes the design and implementation of \Sys's distributed runtime system (\S\ref{subsec:design} and \S\ref{subsec:implementation}), analyzes the computation and memory overheads of speculation and verification (\S\ref{subsec:overhead}), and introduces potential LLM applications that can benefit from \Sys's techniques (\S\ref{subsec:application}).

\subsection{\Sys's Runtime Design}
\label{subsec:design}
\Cref{fig:workflow} shows the workflow for one iteration of speculative inference and verification.
\Sys's {\em request manager} receives LLM serving requests and schedules these requests for serving by adapting the {\em iteration-level scheduling} policy from Orca~\cite{yu2022orca}. Specifically, \Sys iteratively selects requests from a pool of pending requests and performs one iteration of speculative inference and token tree verification for the selected requests. Since SSMs are small and can fit in one GPU, \Sys equally distributes GPUs across SSMs and serves these SSMs using data parallelism. 
For example, \Cref{fig:workflow} shows how \Sys serves two SSMs and four requests (i.e., $r_1$, $r_2$, $r_3$, and $r_4$) on four GPUs.
The SSM-generated tokens are sent back to the request manager, which produces a speculated token tree for each request using the tree merge algorithm introduced in \S\ref{sec:verification}.

\Sys serves an LLM using the hybrid parallelization strategy introduced in Megatron-LM~\cite{shoeybi2020megatronlm}, which uses tensor model parallelism for parallelizing each Transformer layer across GPUs within a node, and uses pipeline model parallelism for partitioning Transformer layers across nodes. 
All GPUs perform the tree-based parallel decoding (see \S\ref{subsec:optimization}) to compute tree attention scores and send the LLM-generated tokens back to the request manager, which finally verifies the speculated tokens against the LLM's output (see \S\ref{subsec:verification}). 

Note that the overhead introduced by the request manager (i.e., request scheduling, token tree merge, and verification) is negligible compared to the execution time of LLM inference. In addition, \sys's request manager and GPU workers only communicate tokens and do not transfer the vector representations of these tokens, which again introduces negligible communication overheads.


\paragraph{Continuous batching.}
\Sys uses {\em continuous batching} introduced in Orca~\cite{yu2022orca} to serve multiple LLM inference requests in parallel. Specifically, \sys schedules LLM execution at the granularity of iterations instead of requests. After each LLM decoding iteration, \sys checks each request's status and sends the generated results of all finished requests to the client. This design also allows \sys to start processing newly arrived requests without waiting for all current requests to complete.

\subsection{\Sys's Implementation}
\label{subsec:implementation}
\Sys was implemented on top of FlexFlow~\cite{jia2019flexflow, unger2022unity}, a distributed multi-GPU runtime for DNN computation. FlexFlow exposes an API that allows users to define a DNN model in terms of its layers.
\rev{It is compatible with PyTorch’s model definition due to the alignment of underlying operators. For example, the open-source LLMs from HuggingFace~\cite{huggingface} can be directly imported into \sys for serving without modification.}
Users can also provide a parallelization strategy, specifying the degree of data, model, and pipeline parallelism for each layer. 
%
A DNN is represented as a computational graph where each node is a region of memory, and each edge is an operation on one or more regions. Operations can be represented using three levels of abstraction: layers, operators, and tasks. The FlexFlow compiler transforms the computational graph from the highest abstractions (i.e., layers) to the lowest (i.e., tasks). Tasks are also the unit of parallelization; they are non-preemptible, and are executed asynchronously. 

\rev{
\paragraph{CUDA kernel optimizations.}
Directly launching cuBLAS and cuDNN kernels for calculating attention results in high kernel launch overhead and does not leverage the shared memory available on modern GPUs. To address this inefficiency, \Sys uses a customized kernel built on top of FasterTransformer~\cite{fastertransformer} for computing attention.
Within this kernel, each thread block computes a single head for a single request. The process begins with loading the query tensor into GPU shared memory accessible by all threads within a thread block. Each thread then performs a segment of the query/key product and broadcasts the result to other threads for computing the max query/key product and exponential sum.
To support tree-based parallel decoding, \sys computes all tokens within a tree in parallel and leverages the topology-aware causal mask to preserve casuality.}

\subsection{Overhead of Speculation and Verification}
\label{subsec:overhead}
\Sys accelerates generative LLM inference at the cost of additional memory and computation overheads. This section analyzes these overheads and shows that they are generally one or two orders of magnitude smaller than the memory and computation cost of executing LLM inference.

\paragraph{Memory overhead.} The memory overhead of \Sys's speculation-verification approach comes from two aspects. First, in addition to serving an LLM, \Sys also needs to allocate memory for saving the parameters of one or multiple SSMs, which collectively speculate the LLM's output. 
Our evaluation shows that \Sys can achieve significant performance improvement by using SSMs 100-1000$\times$ smaller than the LLM. As a result, hosting each SSM increases the overall memory requirement by less than 1\%. 
A second source of memory overhead comes from the token tree verification engine, which verifies an entire token tree instead of decoding a single token. 
Therefore, additional memory is needed for caching the keys and values, and storing the attention scores for all tokens. 
Due to the necessity for supporting very long sequence length in today's LLM serving, we observe that the memory overhead associated with token tree is negligible compared to key-value cache.

\paragraph{Computation overhead.} Similarly, the computation overhead introduced by speculation and verification also comes from two aspects.
First, \Sys needs to run SSMs in the incremental-decoding mode to generate candidate tokens. When multiple SSMs are employed, \sys processes these SSMs in parallel across GPUs to minimize speculation latency.
Second, \Sys verifies a token tree by computing the attention outputs for an entire token tree, most of which do not match the LLM's output and therefore are unnecessary in the incremental-decoding inference.
However, the key-value cache mechanism of existing LLM inference systems prevents them from serving a large number of requests in parallel, resulting in under-utilized computation resources on GPUs when serving LLMs in incremental decoding. \Sys's token tree verification leverages these under-utilized resources and therefore introduces negligible runtime overhead compared to incremental decoding.

\subsection{Applications}
\label{subsec:application}
Our speculative inference and token tree verification techniques can be directly applied to a variety of LLM applications. We identify two practical scenarios where LLM inference can significantly benefit from our techniques.

\paragraph{Distributed LLM inference.} 
The memory requirements of modern LLMs exceed the capacity of a single compute node with one or multiple GPUs, and the current approach to addressing the high memory requirement is distributing the LLM's parameters across multiple GPUs~\cite{miao2023spotserve}.
For example, serving a single inference pipeline for GPT-3 with 175 billion parameters requires more than 16 NVIDIA A100-40GB GPUs to store the model parameters in single-precision floating points.
Distributed LLM inference is largely limited by the latency to transfer intermediate activations between GPUs for each LLM decoding step. 
While \Sys's approach does not directly reduce the amount of inter-GPU communications, its verification mechanism can increase the communication granularity and reduce the number of decoding steps.

\paragraph{Offloading-based LLM inference.}
Another practical scenario that can benefit from \Sys's techniques is offloading-based LLM inference, which leverages 
CPU DRAM to store an LLM's parameters and loads a subset of these parameters to GPUs for computation in a pipeline fashion~\cite{sheng2023highthroughput}.
By opportunistically verifying multiple tokens, \Sys can reduce the number of LLM decoding steps and the overall communication between CPU DRAM and GPU HBM.
\section{Evaluation}
\label{sec:eval}

\begin{figure*}
    \centering
    \includegraphics[scale=0.52]{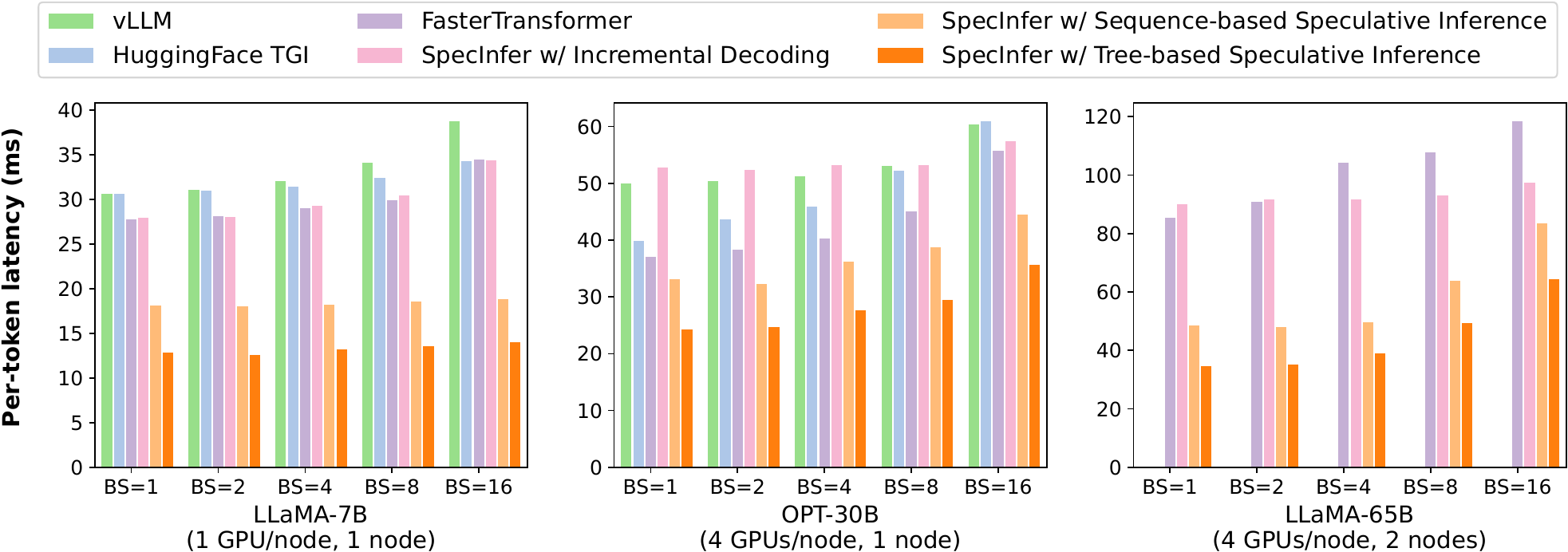}
    \caption{\rev{Comparing the end-to-end inference latency of \Sys with existing systems. Numbers in parenthesis show the number of GPUs and compute node used to serve each LLM. All systems parallelize LLM inference by combining tensor model parallelism (within a node) and pipeline parallelism (across nodes).}
    }
    \label{fig:latency}
\end{figure*}

\subsection{Experimental Setup}
\label{subsec:setup}

\paragraph{LLMs.} To compare the runtime performance of \Sys with existing LLM serving systems, we evaluate these systems using two publicly available LLM families: OPT~\cite{zhang2022opt} and \llama~\cite{touvron2023llama}. 
More specifically, we select \llama-7B, OPT-13B, OPT-30B, and \llama-65B as the LLMs, and \llama-68M and OPT-125M as the SSMs.
The pre-trained model parameters for the LLMs and SSMs were obtained from their HuggingFace repositories~\cite{huggingface}, and we describe how \Sys collectively boost-tunes multiple SSMs in~\cite{miao2023specinfer}.

\paragraph{Datasets.} We evaluate \Sys on five datasets: Chatbot Instruction Prompts (CIP)~\cite{chatbot2023}, ChatGPT Prompts (CP)~\cite{mohamed2023chatgpt}, WebQA~\cite{berant-etal-2013-semantic}, Alpaca~\cite{alpaca, peng2023instruction}, and PIQA~\cite{Bisk2020}. 
We only use the prompts/questions from these datasets to form our input prompts to simulate real-world conversation traces.

\paragraph{Platform.} The experiments were conducted on two AWS {\tt g5.12xlarge} instances, each of which is equipped with four NVIDIA A10 24GB GPUs, 48 CPU cores, and 192 GB DRAM. Nodes are connected by 100 Gbps Ethernet.

Our experiments use the expansion-based method (see \Cref{sec:speculation}) for constructing token trees and use the expansion configuration $\langle 1, 1, 3, 1, 1, 1, 1, 1\rangle$, which provides good results for our benchmarks. We analyze the impact of expansion configurations in \S\ref{subsec:eval_token_tree}, evaluate tree-based parallel decoding and multi-step speculative sampling in \S\ref{subsec:eval_tree_based_decoding} and \S\ref{subsec:eval_msss}, and finally compares the expansion- and merge-based tree construction methods in~\cite{miao2023specinfer}.


\subsection{Distributed LLM Inference}
\label{subsec:eval_distributed}
We compare the end-to-end distributed LLM inference performance among \Sys, vLLM~\cite{vllm}, HuggingFace Text Generation Inference (TGI)~\cite{huggingface_tgi}, and FasterTransformer~\cite{fastertransformer} on \llama-7B, OPT-30B, and \llama-65B.
For \llama-7B and OPT-30B, all systems serve the two LLMs in half-precision floating points across one and four A10 GPUs using tensor model parallelism.
\llama-65B do not fit on four GPUs on a single node, therefore both FasterTransformer and \Sys serve it on eight A10 GPUs on two nodes by combining tensor model parallelism within each node and pipeline model parallelism across nodes.
vLLM and HuggingFace TGI do not support pipeline model parallelism and cannot serve an LLM on multiple nodes.

To rule out potential effects of our system implementation, we also evaluate \Sys with two additional configurations. First, \sys with {\em incremental decoding} evaluates the runtime performance of our implementation when the speculator generates empty token trees, and the verifier verifies exactly one token in each decoding step.
Second, \sys with {\em sequence-based speculative inference} serves as a reference for existing speculative inference system and is enabled by using a single pre-trained SSM and sequence-based decoding.

We use prompts from the five datasets described in \S\ref{subsec:setup}. For each prompt, we let all systems generate up to 128 new tokens and report the average per-token latency in \Cref{fig:latency}.
Note that \Sys may generate more than 128 new tokens since the verifier can verify multiple tokens in each iteration. 
In this case, we truncate \Sys's output to 128 tokens.
\Sys with incremental decoding achieves on-par performance as existing  systems. This is because all systems use the same strategies to parallelize LLM inference across GPUs and use the same kernel libraries (i.e., cuDNN, cuBLAS, and cuTLASS) to execute inference computation on GPUs.
With tree-based speculative inference and verification, \Sys outperforms incremental decoding systems by 1.5-2.5$\times$ for single-node, multi-GPU inference and by 2.4-2.8$\times$ for multi-node, multi-GPU inference, while generating the exact same sequence of tokens as incremental decoding for all prompts.
\rev{The speedup comes from leveraging spare GPU resources to perform tree-based parallel decoding while maintaining the same per-iteration latency as incremental decoding.}

\rev{Compared to sequence-based speculative inference, \sys's tree-based approach further reduces LLM serving latency by 1.2-1.5$\times$. The improvement is achieved by (1) leveraging token trees to optimize speculative performance, (2) using tree-based parallel decoding to verify an entire token tree in parallel, and (3) performing multi-step speculative sampling to improve verification performance. We further evaluates these aspects in \S\ref{subsec:eval_token_tree}, \S\ref{subsec:eval_tree_based_decoding}, and \S\ref{subsec:eval_msss}.}

\iffinal
Note that \sys's performance improvement over existing systems reduces as the batch size (i.e., number of concurrent requests) increases.
This is because \sys leverages spare GPU resources to perform tree-based parallel decoding while maintaining the same per-iteration latency as incremental decoding.
A larger batch size introduces more parallelizable computation for incremental decoding, and thus less spare GPU resources that can be leveraged by \sys.
On the flip side, larger batch sizes also increase the end-to-end latency of each request, as shown in \Cref{fig:latency}.
Overall, \sys is most beneficial for {\em low-latency} LLM inference.
\fi 


\subsection{Offloading-based LLM Inference}
\label{subsec:eval_offload}
\begin{figure}
    \centering
    \includegraphics[scale=0.33]{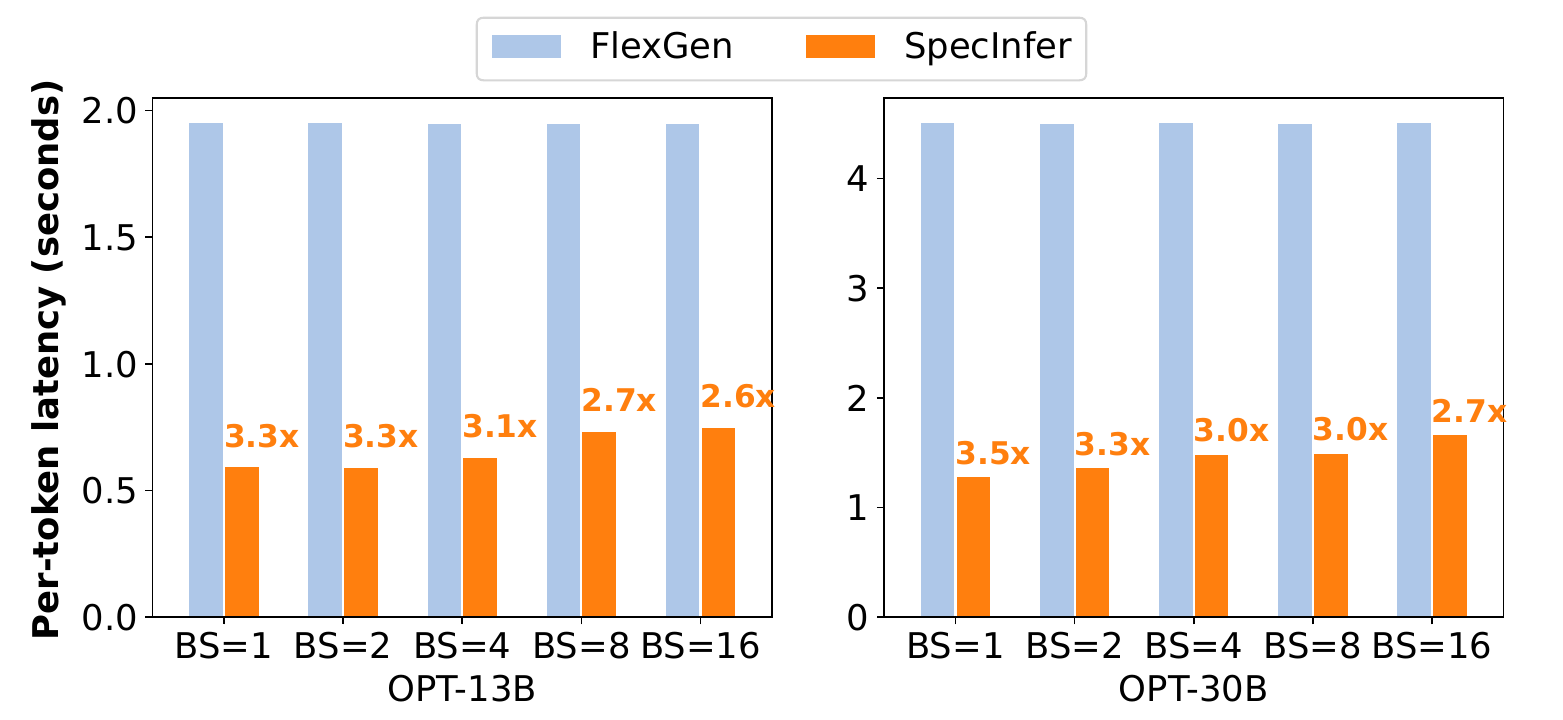}
    \caption{Comparing the end-to-end offloading-based inference latency of FlexGen and \Sys. Both FlexGen and \Sys perform model offloading to serve OPT-13B and OPT-30B models on a single 24GB A10 GPU.}
    \label{fig:eval_offload}
\end{figure}

Another important application of \Sys is offloading-based LLM inference, where the system offloads an LLM's parameters to CPU DRAM and loads a subset of these parameters to GPUs for inference computation in a pipeline fashion.
We compare the end-to-end offloading-based LLM inference performance between \Sys and FlexGen~\cite{sheng2023flexgen} using a single 24GB A10 GPU and two LLMs (i.e., OPT-13B and OPT-30B), both of which exceed the memory capacity of an A10 GPU and requires offloading for serving. Both \Sys and FlexGen retain all model parameters on CPU DRAM. During computation, the demand weights are loaded from the CPU to the GPU.
\Cref{fig:eval_offload} shows the results. Compared to FlexGen, \Sys reduces the per-token latency by 2.6-3.5$\times$. 
Since offloading-based LLM inference is mostly bottlenecked by the communication between CPU DRAM and GPU HBM for loading an LLM's parameters, \Sys's improvement over existing systems is achieved by opportunistically verifying multiple tokens, which in turn reduces the number of LLM decoding steps and data transfers between CPU and GPU.

\subsection{\rev{Token Tree Construction}}
\label{subsec:eval_token_tree}

\begin{figure}
    \centering
    \includegraphics[width=\linewidth]{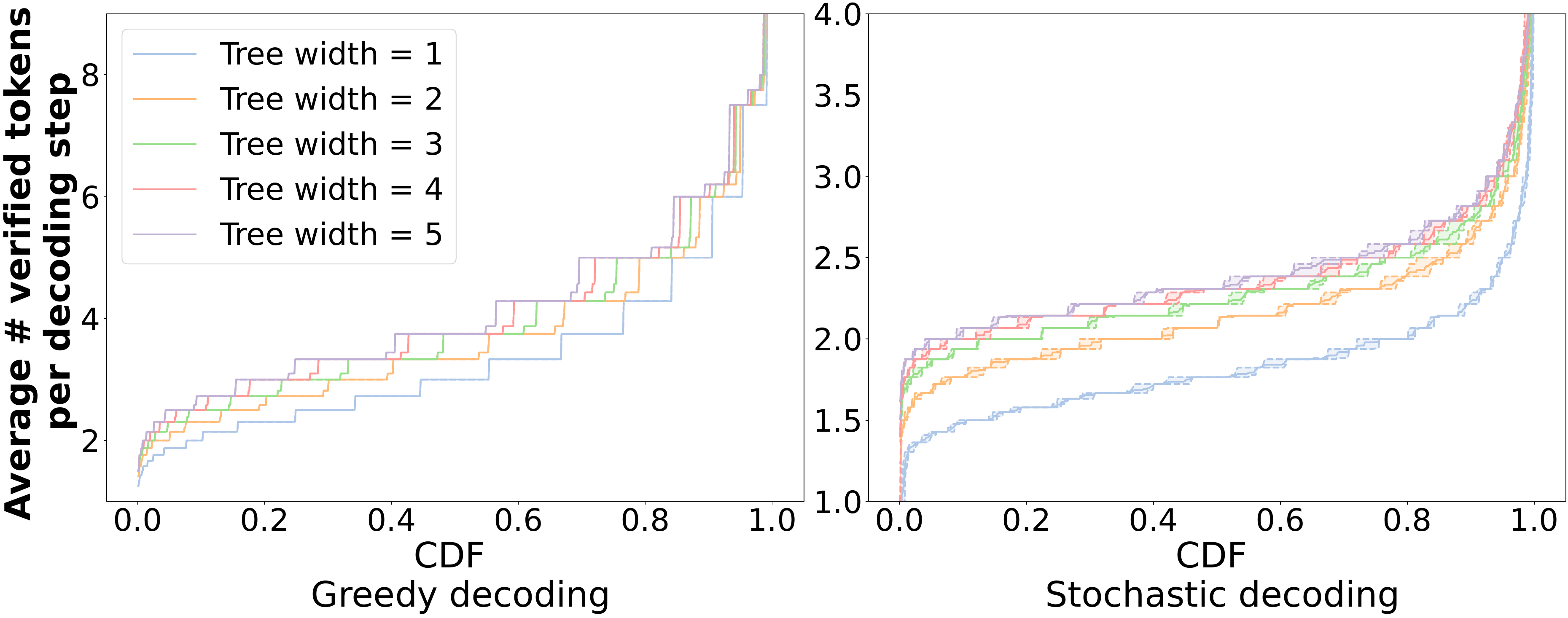}
    \caption{\rev{Comparing speculative performance of \sys with different token tree structures.}}
    \label{fig:eval_cdf}
\end{figure}

\rev{This section evaluates the expansion-based token tree construction mechanism. We first study how token tree width affects \Sys's speculative performance. In this experiment, we use \llama-7B and \llama-68M as the LLM and SSM, and use the expansion configuration $\langle 1, 1, k, 1, 1, 1, 1, 1\rangle$ (i.e., expanding at the third token), where $k$ is the token tree width.
\Cref{fig:eval_cdf} shows the cumulative distribution function (CDF) of the average number of verified tokens per decoding step for all prompts in the Alpaca dataset~\cite{alpaca}.
Compared to sequence-based speculation (i.e., tree width = 1), leveraging token trees can reduce LLM decoding steps by 1.2-1.5$\times$ for greedy decoding and by 1.3-1.4$\times$ for stochastic decoding.}

\begin{table}
\begin{center}
\caption{\rev{Average number of tokens verified by \sys in a decoding step. We use \llama-7B and \llama-68M as the LLM and SSM, and use different tree widths for constructing a token tree. The speculation length is 8.}}
\label{tab:cbt-llama}

\begin{tabular}{ll|ccccc}
\toprule
& & \multicolumn{5}{c}{{\bf Token tree width}} \\
& Dataset   & 1 & 2 & 3 & 4 & 5 \\ \midrule
\multirow{5}{*}{\rotatebox[origin=c]{90}{\parbox{2cm}{\bf \centering Greedy\\ decoding}}}
& Alpaca     & 2.95     & 3.07    & 3.21     & 3.33     & \textbf{3.43}     \\
& CP         & 2.58     & 3.24    & 3.46     & 3.59     & \textbf{3.69}     \\
& WebQA      & 2.27     & 2.69    & 2.86     & 2.98     & \textbf{3.07}     \\ 
& CIP        & 2.73     & 3.40    & 3.62     & 3.79     & \textbf{3.91}     \\
& PIQA       & 2.18     & 2.80    & 2.97     & 3.10     & \textbf{3.21}     \\  
\midrule
\multirow{5}{*}{\rotatebox[origin=c]{90}{\parbox{2cm}{\bf \centering Stochastic\\ decoding}}}
& Alpaca     & 1.79     & 2.11    & 2.26     & 2.32     & \textbf{2.38}     \\
& CP         & 1.69     & 1.99    & 2.15     & 2.23     & \textbf{2.28}     \\
& WebQA      & 1.64     & 1.93    & 2.08     & 2.15     & \textbf{2.21}     \\
& CIP        & 1.72     & 2.05    & 2.19     & 2.28     & \textbf{2.29}     \\
& PIQA       & 1.67     & 1.93    & 2.08     & 2.15     & \textbf{2.21}         \\ 
\bottomrule
\end{tabular}
\end{center}
\end{table}

\begin{figure}
    \centering
    \includegraphics[scale=0.45]{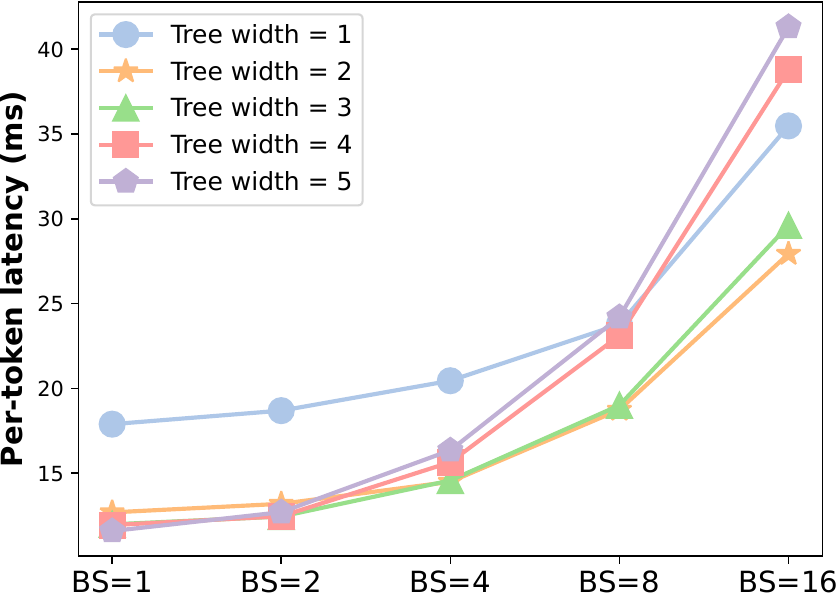}
    \caption{\rev{\Sys's end-to-end inference latency with different tree widths. We use \llama-7B and \llama-68M as the LLM and SSM.}}
    \label{fig:eval_tree_branches}
\end{figure}

\rev{A larger token tree width reduces the LLM decoding steps to process a request at the cost of increased verification overhead, since \sys must verify more tokens.
\Cref{fig:eval_tree_branches} compares the end-to-end inference latency of \sys using different tree widths.
For small batch sizes (i.e., BS = 1 and 2), using a large tree width can consistently reduce per-token latency, since \sys can leverage sparse GPU resources to verify more tokens in parallel while maintaining the same per-iteration latency.
For large batch sizes (i.e., BS $\geq 4$), using a large tree width increases the latency to verify a token tree due to less sparse GPU resources that can be leveraged by \sys, and a tree width of $2$ or $3$ achieves the best performance by striking a perfect balance between speculative performance and verification latency.}

\subsection{\rev{Tree-based Parallel Decoding}}
\label{subsec:eval_tree_based_decoding}

\begin{figure}
    \centering
    \includegraphics[scale=0.45]{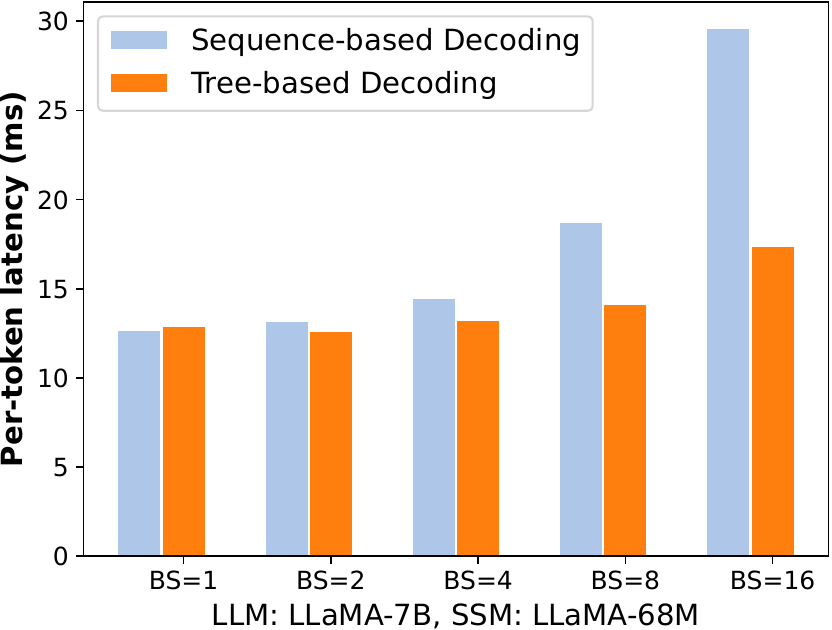}
    \caption{\rev{Comparing \Sys's tree-based parallel decoding with the sequence-based decoding mechanism employed by existing LLM inference systems.}}
    \label{fig:eval_tree_based_decoding}
\end{figure}

\rev{We now evaluate the effectiveness of \sys's tree-based parallel decoding mechanism, which decodes all tokens of a token tree in parallel.
As a comparison, all existing LLM inference systems use sequence-based decoding, which requires decomposing a token tree into multiple sequences of tokens and processing these sequences using separate resources due to potential key-value cache conflicts (see \S\ref{subsec:optimization}).
%
As shown in \Cref{fig:eval_tree_based_decoding}, \sys's tree-based parallel decoding achieves on-par performance as existing sequence-based decoding mechanism for small batch sizes and outperforms it by up to 1.8$\times$ for large batch sizes. The improvement is realized by (1) eliminating redundant attention computation for sequences with a shared prefix, and (2) fusing tree attention of all tokens in a single kernel through the topology-aware casual mask (see \S\ref{subsec:optimization}).}

\begin{table}
\begin{center}
\caption{\rev{Average number of tokens verified by \sys in a stochastic decoding step with different sampling algorithms. We use \llama-7B and \llama-68M as the LLM and SSM. Each token tree has a width of 5 and a depth of 8.}}
\label{tab:eval_msss}
\resizebox{\linewidth}{!}{%
\begin{tabular}{l|rrr}
\toprule
& {\bf Naive}  & {\bf Multi-Step} & {\bf Improvement} \\ 
& {\bf Sampling} & {\bf Spec. Sampling} & \\ \midrule
Alpaca     & 1.87     & 2.38    & 1.27$\times$\\
CP         & 1.80     & 2.28    & 1.26$\times$\\
WebQA      & 1.73     & 2.21    & 1.28$\times$\\ 
CIP        & 1.79     & 2.29    & 1.28$\times$\\
PIQA       & 1.73     & 2.21    & 1.28$\times$\\  
\bottomrule
\end{tabular}%
}
\end{center}
\end{table}

\subsection{\rev{Multi-Step Speculative Sampling}}
\label{subsec:eval_msss}

\rev{This section evaluates how our multi-step speculative sampling (MSS) and the \textproc{VerifyStochastic} algorithm improves the speculative performance of \sys when performing stochastic decoding.
We use naive sampling as a baseline where \sys directly samples the next token from the LLM and examines whether the sampled token is included in the speculated token tree (see \S\ref{subsec:verification}).
Since different sampling algorithms involve identical speculation and verification overheads, we focus on the average number of tokens that can be verified in each stochastic decoding step in this experiment.
\Cref{tab:eval_msss} shows the results. Compared to naive sampling, MSS can consistently improve the number of verified tokens by 1.2-1.3$\times$ on average across a variety of prompt datasets, while guaranteeing the same output distribution with the LLM.
}
\section{Related Work}
\label{sec:related}

Transformer-based LLMs have demonstrated significant potential in numerous human-level language modeling tasks by continuously increasing their sizes~\cite{vaswani2017attention, rae2021scaling, smith2022using, du2022glam, chowdhery2022palm}.
As GPT-3 becomes the first model to surpass 100B parameters~\cite{brown2020language}, multiple LLMs ($>$100B) have been released, including OPT-175B~\cite{zhang2022opt}, Bloom-176B~\cite{scao2022bloom}, and PaLM~\cite{chowdhery2022palm}.
Recent work has proposed a variety of approaches to accelerating generative LLM inference, which can be categorized into two classes.

\paragraph{Lossless acceleration.}
Prior work has explored the idea of using an LLM as a verifier instead of a decoder to boost inference.
For example, Yang et al.~\cite{yang2023inference} introduced {\em inference with reference}, which leverages the overlap between an LLM's output and the references obtained by retrieving documents, and checks each reference's appropriateness by examining the decoding results of the LLM.
Motivated by the idea of speculative execution in processor optimizations~\cite{burton1985speculative, hennessy2011computer}, recent work proposed {\em speculative decoding}, which uses a small language model to produce a sequence of tokens and examines the correctness of these tokens using an LLM~\cite{leviathan2022fast, xiaspeculative, stern2018blockwise, chen2023accelerating, gante2023assisted}.
There are two key differences between \Sys and these prior works.
First, instead of only considering a single sequence of tokens, \Sys generates and verifies a token tree, whose nodes each represent a unique token sequence. \Sys performs tree attention to compute the attention output of these token sequences in parallel and uses a novel tree-based decoding algorithm to reuse intermediate results shared across these sequences.
Second, prior attempts generally consider a single small language model for speculation, which cannot align well with an LLM due to the model capacity gap between them.
\rev{\Sys introduces two novel speculation methods, including 1) expanding from a single SSM and 2) merging from multiple fine-tuned SSMs, and the generated token tree largely increases the coverage of the LLM's output.}

Prior work has also introduced a variety of techniques to optimize ML computations on modern hardware platforms. 
For example, TVM~\cite{chen2018tvm} and Ansor~\cite{zheng2020ansor} automatically generate kernels for a given tensor program. TASO~\cite{jia2019taso} and PET~\cite{wang2021pet} automatically discover graph-level transformations to optimize the computation graph of a DNN.
\Sys's techniques are orthogonal and can be combined with these systems to accelerate generative LLM computation, which we believe is a promising avenue for future work.


\paragraph{Lossy acceleration.}
BiLD~\cite{biglittletransformer} is a speculative decoding framework that uses a single SSM to accelerate LLM decoding. Unlike the systems mentioned above, the acceleration is lossy: speed comes at the cost of a possible degradation in the generated tokens. 
Another line of research leverages model compression to reduce LLM inference latency while compromising the predictive performance of the LLM.
For example, prior work proposed to leverage weight/activation quantization of LLMs to reduce the memory and computation requirements of serving these LLMs~\cite{xiao2022smoothquant, frantar2023gptq, park2022nuqmm, yao2022zeroquant,dettmers2022gpt3}.
Recent work further explores a variety of structured pruning techniques for accelerating Transformer-based architectures~\cite{frantar2023sparsegpt, wang2021spatten, hubara2021accelerated}.
A key difference between \Sys and these prior works is that \Sys does not directly reduce the computation requirement for performing LLM inference, but instead reorganizing LLM inference computation in a more parallelizable way, which reduces memory accesses and inference latency at the cost of manageable memory and computation overheads.

\paragraph{Tree-structured attention.}
Nguyen et al.~\cite{Nguyen2020} introduced {\em tree-structured attention}, a technique that lets a Transformer model capture the hierarchical composition of input text by running the model on the text's parse tree. To process with attention, it uses a one-on-one mapping to encode and decode the tree.
There are two key differences from \sys's tree-based decoding. First, \sys uses a tree to combine candidate sequences to condense prefixes, whereas Nguyen et al. represent a single sequence with its parse tree. \sys does not incorporate parse tree into the LLM, but accelerates inference by verifying decoded sequences in parallel. Second, \sys's attention outputs a token sequence, not a tree.

\paragraph{Multi-sample decoding techniques}
Like tree-based speculative inference, \textit{beam search}, \textit{top-k sampling}, and \textit{top-p sampling} consider multiple candidate token sequences at each step and can prune low-probability options. However, tree-based decoding in \sys speculatively predicts and verifies multiple candidates in parallel against an LLM to reduce decoding iterations and latency, leveraging small speculative models (SSMs). In contrast, beam search and top-k/top-p sampling are decoding strategies applied directly to the LLM's output probabilities to generate high-probability sequences without reducing decoding steps. \sys supports beam search, top-k sampling, and top-p sampling. These techniques are orthogonal decoding optimizations and can be combined with tree-based speculative decoding.

\section{Conclusion}
This paper introduces \Sys, a system that accelerates generative LLM inference with tree-based speculative inference and verification.
A key insight behind \Sys is to simultaneously consider a diversity of speculation candidates to efficiently predict the LLM’s outputs, which are organized as a token tree and verified against the LLM in parallel using a tree-based parallel decoding mechanism.
\Sys significantly reduces the memory accesses to the LLM's parameters and the end-to-end LLM inference latency for both distributed and offloading-based LLM inference.

\iffinal
\section*{Acknowledgement}
We thank Tianqi Chen, Bohan Hou, Hongyi Jin, the anonymous ASPLOS reviewers, and our shepherd Shan Lu for their feedback on this work. This research is partially supported by NSF awards CNS-2147909, CNS-2211882, and CNS-2239351, and research awards from Amazon, Cisco, Google, Meta, Oracle, Qualcomm, and Samsung. 
\fi

\appendix
\section{Artifact Appendix}

\subsection{Abstract}

The artifact contains the code to run \sys, as well as the datasets and scripts that can be used to reproduce the experiments in the paper. 
\subsection{Artifact check-list (meta-information)}


{\small
\begin{itemize}
  \item {\bf Algorithm: } Tree-based Speculative Inference
  \item {\bf Program: } spec\_infer.cc, incr\_decoding.cc
  \item {\bf Compilation: } CMake
  \item {\bf Run-time environment: } CUDA, NCCL, MPI, UCX, Python3.
  \item {\bf Hardware: }  Two AWS g5.12xlarge instances, each with 4 NVIDIA A10 24GB GPUs, 48 CPU cores, and 192 GB DRAM.
  \item {\bf Metrics: } End to-end average latency
  \item {\bf Output: } End-to-end latency
  \item {\bf Experiments: } Server-grade GPU inference, Offloading-based inference
  \item {\bf How much disk space required (approximately)?: } 350GB per node
  \item {\bf How much time is needed to prepare workflow (approximately)?: } 2h
  \item {\bf How much time is needed to complete experiments (approximately)?: } 6h
  \item {\bf Publicly available?: } Yes
  \item {\bf Code licenses (if publicly available)?: } Apache License v2.0
  \item {\bf Data licenses (if publicly available)?: } LLAMA is under the GNU license and OPT is under a Non-commercial license
  \item {\bf Archived (provide DOI)?: }\url{https://doi.org/10.5281/zenodo.10854410}.
\end{itemize}
}

\subsection{Description}

\subsubsection{How to access}
The artifact is released on Github: \url{https://github.com/goliaro/specinfer-ae}. The repository contains \sys's source code, and the instructions to build the framework. We also provide scripts to reproduce the experiments from the paper. To clone the repository, use the following command (make sure to pass the \texttt{--recursive} flag):
\begin{lstlisting}[language=sh]
git clone --recursive \
https://github.com/goliaro/specinfer-ae.git
\end{lstlisting}

\subsubsection{Hardware dependencies}
We run out experiments on two AWS g5.12xlarge instances, each with 4 NVIDIA A10 24GB GPUs, 48 CPU cores, and 192 GB DRAM. We provide instructions to create and setup the instances. 

\subsubsection{Software dependencies}
The following software is required: CUDA 12.1, NCCL, Rust, CMake and Python3. Further, UCX and MPI are required for the multinode experiments. Additional Python dependencies are listed here: \url{https://github.com/flexflow/FlexFlow/blob/inference/requirements.txt}. We recommend using the \texttt{Deep Learning OSS Nvidia Driver AMI GPU PyTorch 2.1.0 (Ubuntu 20.04)} AMI, and provide scripts and a conda environment to install all the remaining dependencies.


\subsubsection{Models}
We use the following LLM/SSM models for our experiments (for each model, we specify in parentheses the corresponding HuggingFace repository): LLaMA-68M (JackFram/llama-68m), LLaMA-7B (huggyllama/llama-7b), LLaMA-65B (huggyllama/llama-65b), OPT-125M (facebook/opt-125m), OPT-13B (facebook/opt-13b), OPT-125M (facebook/opt-30b). You can download all these models with the script: 
\begin{lstlisting}[language=sh]
./download_models.sh
\end{lstlisting}

\subsection{Installation}


To reproduce the experiments, you will need access to two AWS g5.12xlarge instances (or other machines with the same GPU/CPU/network specs). If you are using the preconfigured instances we provided, you can skip this step. 
\paragraph{Launching the instances}
Launch two AWS g5.12xlarge instances using the \texttt{Deep Learning OSS Nvidia Driver AMI GPU PyTorch 2.1.0 (Ubuntu 20.04)} AMI. Make sure to place the instances in a \href{https://docs.aws.amazon.com/AWSEC2/latest/UserGuide/placement-groups.html}{placement group} that utilizes the cluster strategy to achieve low-latency network performance. Attach the same security group to all instances and add an inbound rule in the security group to allow all incoming traffic from the same security group. For example, you can add the following rule: Type: All TCP, Source: Anywhere-IPv4.

\paragraph{Installing the prerequisites}
After gaining access to the AWS instances, install the prerequisites by following the steps below. First, activate the conda shell support by running \texttt{conda init bash}, and then restarting the shell session.
Next, create the conda environment with all the required dependencies by running:
\begin{lstlisting}[language=sh]
conda env create -f FlexFlow/conda/flexflow.yml
conda activate flexflow
\end{lstlisting}

\paragraph{Multinode setup}
Download and build UCX by running the \texttt{install\_ucx.sh} script. Next, if you are running \sys on two AWS instances, you will need to configure MPI so that the two instances are mutually accessible. Pick a main node, and create a SSH key pair with:
\begin{lstlisting}[language=sh]
    ssh-keygen -t ed25519
\end{lstlisting}
Append the contents of the public key (\texttt{\textasciitilde/.ssh/id\_ed25519.pub}) to the \texttt{\textasciitilde/.ssh/authorized\_keys} file on BOTH the main and secondary machine. Note that if the \texttt{.ssh} folder or the \texttt{authorized\_keys} file do not exist, you will need to create them manually. 
Finally, create a file at the path \texttt{\textasciitilde/hostfile} with the following contents:
\begin{lstlisting}[language=sh]
<main_node_private_ip> slots=4
<secondary_node_private_ip> slots=4
\end{lstlisting}
replacing <main\_node\_private\_ip> and <secondary\_node\_private\_ip> with the private IP addresses of the two machines, and the number of slots with the number of GPUs available (if you are using the recommended AWS instances, you will use a value of 4). You can find each machine's private IP address by running the command (and use the first IP value that is printed):
\begin{lstlisting}[language=sh]
    hostname -I
\end{lstlisting}

\paragraph{Install \sys}

To install \sys, run the script: 
\begin{lstlisting}[language=sh]
./install_specinfer.sh
\end{lstlisting}

\subsection{Basic Test}
To ensure that \sys is installed correctly and is functional, run the \texttt{basic\_test.sh} script. This script will test the basic incremental decoding and speculative inference functionalities, on both single and multi nodes. It will also test the support for offloading. The test passes if it prints the "Test passed!" message. 

\subsection{Experiment workflow}
The artifact comes with scripts to gather the data that can be used to reproduce the results from the paper. It also comes with scripts that can be used to convert the output data into CSV format for plotting. 

\paragraph{Running Experiments}
We run the following two experiments to evaluate \sys under different hardware setups. The output data will be saved to the \texttt{FlexFlow/inference/output} path. 

\begin{itemize}
    \item \textbf{Server-grade GPU evaluation.} This experiment tests the performance of \sys on server-grade GPUs. The LLMs and SSMs are loaded in GPU memory, and we measure the end-to-end inference latency using 1 node, and 2 nodes. In the single node case, we measure the performance using 1 GPU, or 4 GPUs. In the multinode case, we use 4GPUs per node. The experiments use LLAMA-7B, OPT-30B and LLAMA-65B as the LLMs, and LLAMA-68M and OPT-125M as SSMs. The experiment runs \sys in three different modes: incremental decoding, sequence-based speculative decoding, and tree-based speculative decoding. The former two are used to obtain data for the ablation study, and the latter is the novel inference mode proposed by \sys, and will be deployed by the user. To run the server-grade GPU evaluation, run:
\begin{lstlisting}[language=sh]
./server_gpu_experiments.sh
\end{lstlisting}
\item \textbf{Offloading evaluation.} This experiment tests the performance of \sys when loading only a subset of parameters in GPU memory, while offloading the remaining ones on CPU DRAM. This technique is used to perform inference when the target model is larger than the available GPU memory. In the experiment, \sys uses a single GPU and swaps the model's weights to and from the CPU. To run the offloading evaluation, run: 
\begin{lstlisting}[language=sh]
./offloading_experiments.sh
\end{lstlisting}
\end{itemize}

\paragraph{Third-party frameworks}
Please follow the vLLM, FasterTransformer, and HuggingFace TGI, and FlexGen official documentation to reproduce the performance of the third-party frameworks under the experiment scenarios.

\paragraph{Output data}
The scripts above will generate data at the \texttt{FlexFlow/inference/output} path. For each scenario, a \texttt{.txt} file contains the generated output for each prompt, and a \texttt{.out} file contains the stdout logs. The quality of the generated output can be evaluated visually and compared with the output from third-party inference frameworks.
We provide scripts to parse the raw output data and generate CSV files that can be used to generate the paper's figures. The README provides all details on the scripts and the mapping between CSV files and figures.

\subsection{Evaluation and expected results}
The data from the CSV files should show similar performance to the figures from the paper. Some variability is to be expected, but overall, \sys should behave according to Figures 7-11 from the paper. 

\subsection{Experiment customization}
Users can edit the configuration parameters from the evaluation scripts to change various parameters, such as the number of GPUs/CPUs, GPU/CPU memory, batch size, LLM/SSM models used, prompt dataset, full vs. half-precision, and the maximum number of tokens to generate.


\balance
\bibliographystyle{plain}
\bibliography{reference} 


\end{document}